\documentclass{IEEEtran}

\ifCLASSOPTIONcompsoc
  \usepackage[nocompress]{cite}
\else
  \usepackage{cite}
\fi
\ifCLASSINFOpdf
\else
\fi
\newcommand\MYhyperrefoptions{bookmarks=true,bookmarksnumbered=true,
pdfpagemode={UseOutlines},plainpages=false,pdfpagelabels=true,
colorlinks=true,linkcolor={green},citecolor={blue},urlcolor={black},backref=true,
pdftitle={Training Segmentation on Heterogeneous Datasets},
pdfsubject={Deep Learning},
pdfauthor={Panagiotis Meletis},
pdfkeywords={semantic segmentation, heterogeneous datasets, multi-dataset training, multi-domain training, weakly-supervised learning, semi-supervised learning}}
\ifCLASSINFOpdf
\usepackage[\MYhyperrefoptions,pdftex]{hyperref}
\else
\usepackage[\MYhyperrefoptions,breaklinks=true,dvips]{hyperref}
\usepackage{breakurl}
\fi

\usepackage{amsmath}
\usepackage{amsfonts}
\usepackage{amssymb}

\usepackage{comment}

\usepackage{makecell}

\usepackage{pifont}
\usepackage[x11names, svgnames]{xcolor}

\usepackage{floatrow}
\newfloatcommand{capbtabbox}{table}[][\FBwidth]

\newcommand*\rotnighty{\rotatebox[origin=IB]{90}}
\newcommand*{\multirowrot}[2]{\multirow[b]{#1}{*}{\rotatebox[origin=IB]{90}{#2}}}
\newcommand*{\mrowrot}[2]{\multirow[b]{#1}{*}{\rotatebox[origin=IB]{90}{#2}}}
\newcommand*{\mrowcellrot}[3]{\multirow[b]{#1}{*}{\rotatebox[origin=IB]{90}{\makecell[#2]{#3}}}}

\usepackage{multirow}

\usepackage{subcaption}

\usepackage{url}

\usepackage{bm}
\newcommand{\bsigma}{\bm{\sigma}}

\usepackage{xspace}
\makeatletter
\DeclareRobustCommand\onedot{\futurelet\@let@token\@onedot}
\def\@onedot{\ifx\@let@token.\else.\null\fi\xspace}
\def\eg{\emph{e.g}\onedot}

\def\ie{\emph{i.e}\onedot}

\makeatother

\usepackage{mathtools}
\DeclarePairedDelimiter{\abs}{\lvert}{\rvert}

\usepackage{makecell}

\usepackage{stmaryrd}

\DeclareMathOperator*{\argmax}{arg\,max}

\usepackage{dcolumn}
\newcolumntype{d}[1]{D{.}{.}{#1}}

\usepackage{siunitx}

\usepackage{enumitem}

\usepackage{afterpage}

\def\hlinefull{\noindent\rule{1.0\linewidth}{1pt}}

\usepackage{booktabs}

\usepackage{subfloat}

\newcolumntype{x}[1]{>{\centering\arraybackslash}p{#1pt}}
\newlength\savewidth

\newcolumntype{H}{>{\setbox0=\hbox\bgroup}c<{\egroup}@{}}

\usepackage{orcidlink}

\usepackage[ruled,vlined]{algorithm2e}

\begin{document}
%
\title{Training Semantic Segmentation\\on Heterogeneous Datasets}

%
%
%
%

\author{
Panagiotis~Meletis\orcidlink{0000-0001-8054-1760}~and~Gijs~Dubbelman\orcidlink{0000-0001-6635-3245}%
\IEEEcompsocitemizethanks{%
	\IEEEcompsocthanksitem P. Meletis (p.c.meletis@tue.nl) and G. Dubbelman (g.dubbelman@tue.nl) are with the Department of Electrical Engineering, Eindhoven University of Technology, Eindhoven, The Netherlands.%
}%
}

\IEEEtitleabstractindextext{%
\begin{abstract}
We explore semantic segmentation beyond the conventional, single-dataset homogeneous training and bring forward the problem of Heterogeneous Training of Semantic Segmentation (HTSS). HTSS involves simultaneous training on multiple heterogeneous datasets,~\ie datasets with conflicting label spaces and different (weak) annotation types from the perspective of semantic segmentation. The HTSS formulation exposes deep networks to a larger and previously unexplored aggregation of information that can potentially enhance semantic segmentation in three directions: i) performance: increased segmentation metrics on seen datasets, ii) generalization: improved segmentation metrics on unseen datasets, and iii) knowledgeability: increased number of recognizable semantic concepts. To research these benefits of HTSS, we propose a unified framework, that incorporates heterogeneous datasets in a single-network training pipeline following the established FCN standard. Our framework first curates heterogeneous datasets to bring them in a common format and then trains a single-backbone FCN on all of them simultaneously. To achieve this, it transforms weak annotations, which are incompatible with semantic segmentation, to per-pixel labels, and hierarchizes their label spaces into a universal taxonomy. The trained HTSS models demonstrate performance and generalization gains over a wide range of datasets and extend the inference label space entailing hundreds of semantic classes.
\end{abstract}

\begin{IEEEkeywords}
semantic segmentation, heterogeneous datasets, multi-dataset/domain training, weakly/semi-supervised training
\end{IEEEkeywords}}

\maketitle

\IEEEdisplaynontitleabstractindextext

%
\IEEEpeerreviewmaketitle

\ifCLASSOPTIONcompsoc
\IEEEraisesectionheading{\section{Introduction}\label{sec:introduction}}
\else
\section{Introduction}
\label{sec:introduction}
\fi

%
%
%

\IEEEPARstart{S}{emantic} Segmentation~\cite{guo2017aro,garcia2018survey,minaee2020image} is a indispensable building block of upstream systems in various domains, such as automated driving~\cite{zhu2017perception,janai2020computer}, biomedical image analysis~\cite{taghanaki2020deep}, virtual/augmented reality, and surveillance~\cite{minaee2020image}. Semantic segmentation is part of the bigger family of image recognition tasks, which include, among others, image classification and object detection. The success of supervised Convolutional Neural Networks (CNNs) in image classification~\cite{krizhevsky2012alexnet,szegedy2015going,he2016deep} set them as the de-facto solution for related image recognition tasks, which is in a large part attributed to the successful utilization of very large datasets. Unlike CNNs trained for classification or detection, where data collection and labeling is easier, CNNs for semantic segmentation face two fundamental data challenges that are analyzed in the next paragraphs.

\begin{figure}
	\centering
	\includegraphics[width=\textwidth]{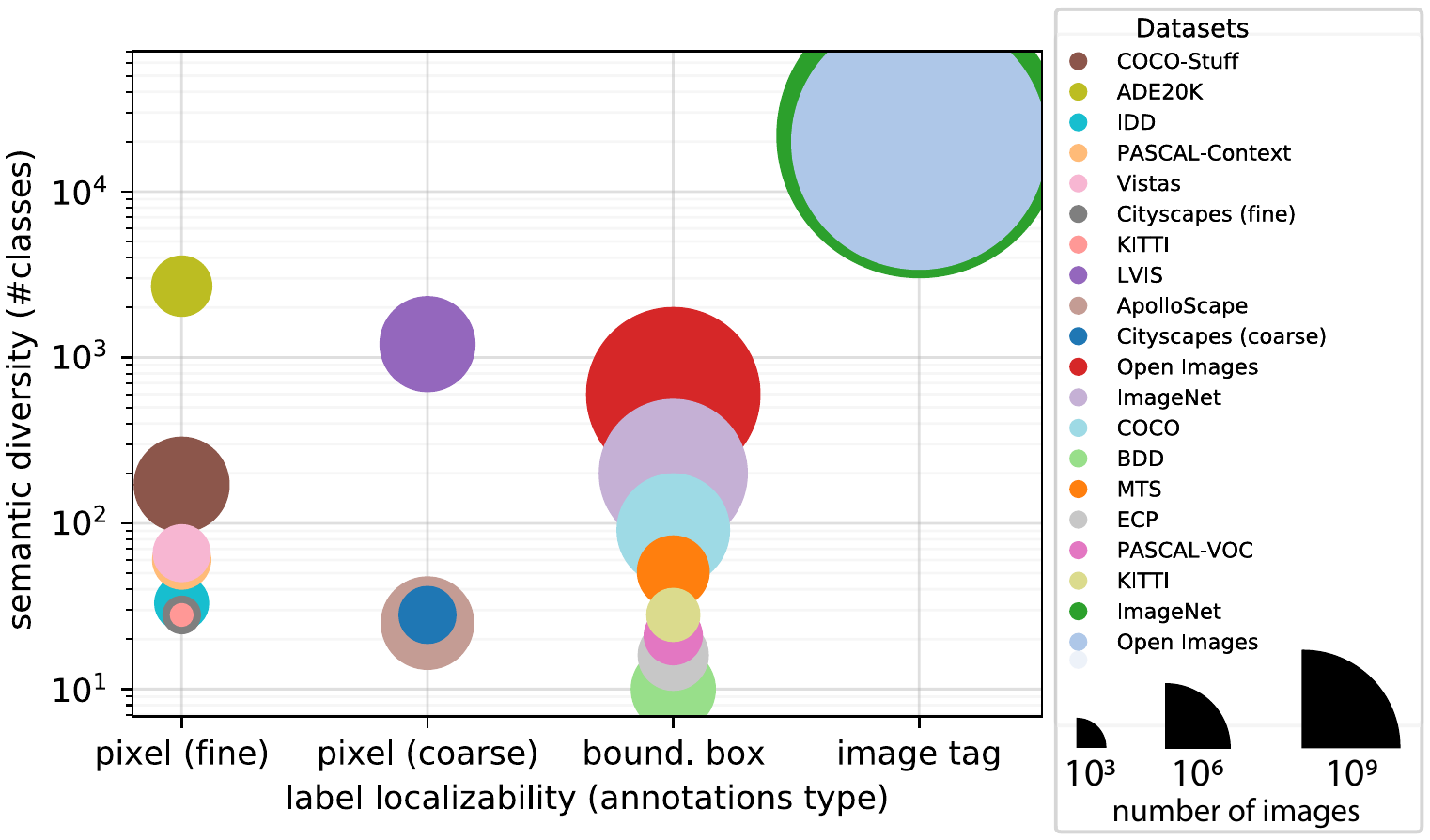}
	\caption{Comparison of various image understanding datasets with respect to their: i) annotation type, ii) number of semantic classes, and iii) number of images (visualized by the area of the circles). Networks for semantic segmentation are typically trained using a single per-pixel (fine) labeled dataset.}
	\label{fig:datasets-statistics}
\end{figure}

\textit{Limited size of existing datasets}.
Datasets for semantic segmentation~\cite{cordts2016Cityscapes,neuhold2017mapillary,varma2019idd} contain typically 100 to 1000 times less annotated images than for image classification and object detection, as visualized in Fig.~\ref{fig:datasets-statistics}. The lack of rich datasets causes CNNs for semantic segmentation to exhibit limited performance when evaluated on seen datasets and poor generalization capabilities on unseen datasets~\cite{zhang2020generalizable,dou2019domain,romijnders2019domain}. This challenge becomes particularly acute in data-scarce areas, such as street-scene understanding. The main reason for the difference in dataset sizes is the required level-of-detail of the annotations for the task at hand. For example, COCO creators~\cite{lin2014microsoft} report that annotating pixels for semantic segmentation of general scenes is 15x slower than drawing bounding boxes, while according to \cite{bearman2016s} annotating pixels is 78x slower than choosing image-level tags.

\textit{Low diversity of represented semantic concepts}. The second fundamental challenge is related to the number of recognizable semantic concepts a CNNs can predict. Typical a CNN trained with segmentation datasets can recognize a few dozens concepts, since fine-grained semantic classes are absent from these datasets. The complexity of (manual) per-pixel labeling practically constrains the annotated semantic classes to represent an order of 100 different scene concepts, while datasets with less spatially-detailed annotations (bounding boxes or tags)~\cite{gupta2019lvis,deng2009imagenet,kuznetsova2020open} can reach up to $1,000 - 10,000$ unique semantic classes (see vert. axis of Fig.~\ref{fig:datasets-statistics}).

The straightforward way to address the aforementioned two challenges, is to annotate more images at the pixel level, or refine existing annotations with finer-grained \mbox{(sub-)}classes by manual or semi-automated means. This is a natural yet costly approach, since manual labeling is laborious (\eg 90 min. per image for Cityscapes and Vistas~\cite{neuhold2017mapillary,cordts2016Cityscapes}) and semi-automated procedures result in insufficient quality of annotations when not complemented with human quality control.

These two practical challenges motivate us to research an alternative approach. Having analyzed the annotation quality and semantic diversity of existing image understanding datasets (Fig.~\ref{fig:datasets-statistics}) we observed that, as a whole, they cover thousands of semantic concepts and contain millions of images. Thus, we contemplate a solution that can combine many existing datasets and solve both aforementioned challenges. However, combining datasets is not an easy task, due to their structural difference, but when successful, it allows for training more capable CNN models with minimal extra manual effort.

In this work, we investigate the challenges and benefits of combined dataset training in three directions: segmentation \textit{performance} on seen datasets (validation/testing splits of datasets on which the CNN is trained), \textit{generalization} on unseen datasets (splits of datasets not used during training), and \textit{Knowledgeability},~\ie, the number of recognizable semantic concepts with sufficient segmentation quality. Instead of combining only datasets for semantic segmentation,
~\ie pixel-labeled datasets, a larger candidate pool of heterogeneous datasets from various image understanding tasks is admitted. These datasets, can significantly increase the available training data and the number of semantic classes leading to improvements in all three directions.

Combining different heterogeneous datasets brings up new challenges, since these datasets are created for different tasks or application domains, and thus they contain conflicting or unrelated label spaces and incompatible (weak) annotations. As a consequence, it is impossible to employ established training strategies for semantic segmentation,~\eg a fully convolutional network pipeline (FCN)~\cite{long2015fully}. To advance the state-of-the-art in multi-dataset training, we generalize semantic segmentation training over heterogeneous datasets and analyze the related challenges. Subsequently, we propose a unified methodology that decouples dataset specifics (structure, annotation types, label spaces) from the task formulation of semantic segmentation. In this way, a plethora of existing image understanding datasets can be leveraged under the same consistent and robust FCN-based framework. The contributions of this work can be summarized as follows.
\begin{itemize}[noitemsep,topsep=0pt]
	\item The formulation of the heterogeneous datasets training problem for semantic segmentation (HTSS) and characterization of the challenges.
	\item A methodology for combining label spaces with different semantic granularity (level-of-detail) and with different semantic concepts, thus enabling simultaneous training on datasets with disjoint or conflicting label spaces.
	\item A methodology for consolidating strong (pixel) and weak (bounding-box or image-tag) supervision, and thus enabling simultaneous training with mixed supervision.
	\item The novel \textit{Knowledgeability} metric, which quantifies the number of recognizable semantic classes by a network wrt. achievable performance for these classes, and that can be used to compare the performance of a network across datasets irrespective of the number of classes.
\end{itemize}

\section{Related work}
\label{sec:rel-work}
Multi-dataset training is gaining traction in various areas,~\eg in object detection~\cite{zhou2021simple,zhao2020object}, depth estimation~\cite{ranftl2020depth}, and domain adaptation~\cite{zhao2020multisource,sun2015survey}, since it improves model robustness and generalization capabilities. This work focuses on semantic segmentation and relaxation of the requirements that a dataset has to comply to, in order to be suited for multi-dataset training. The proposed work generalizes related approaches in literature for semantic segmentation~\cite{lambert2020mseg,jain2020scaling}  and complements recent work~\cite{bevandic2020multi},~\cite{ghiasi2021multi},~\cite{wang2022cross},~\cite{redmon2017yolo9000}.

\begin{figure*}
	\centering
	\includegraphics[width=0.9\linewidth]{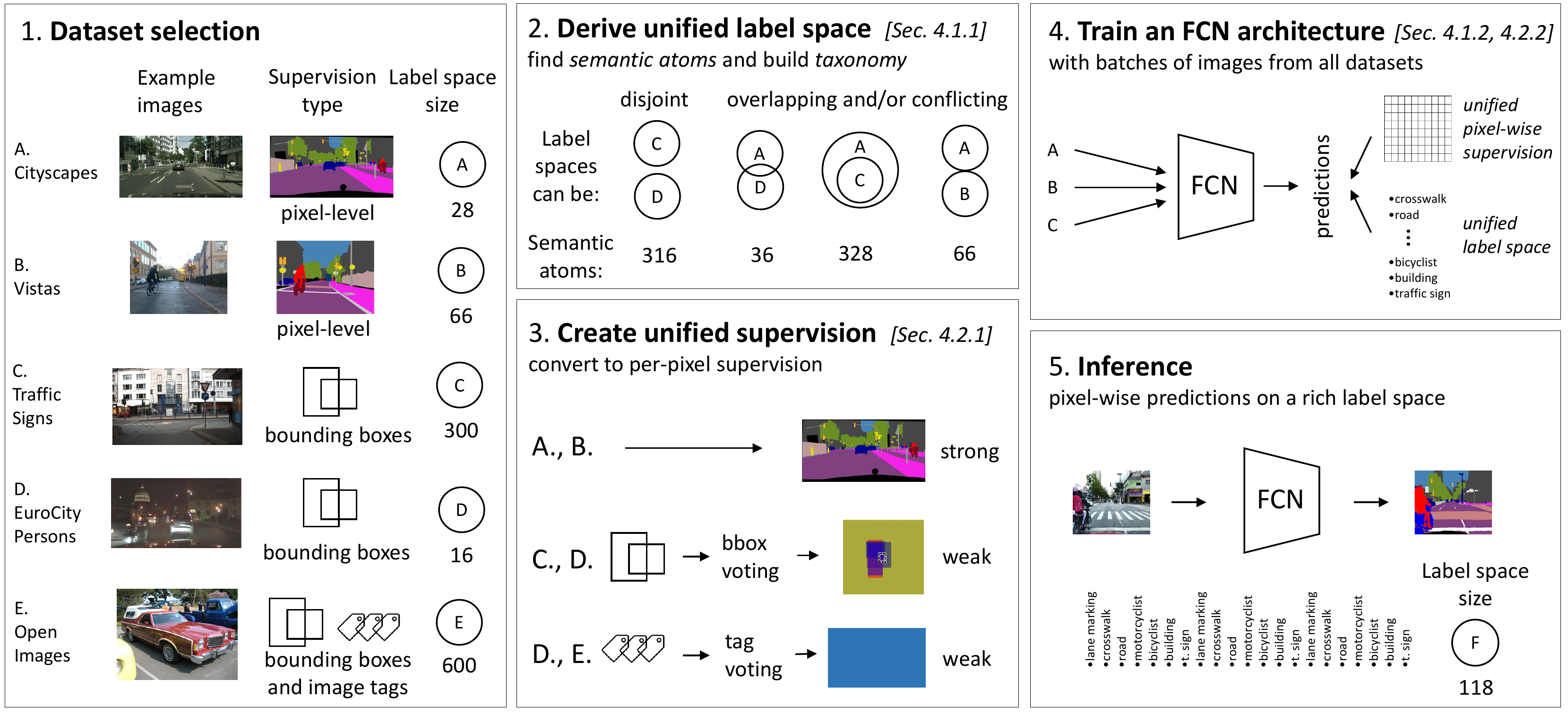}
	\caption{Motivation and overview of the proposed framework. HTSS aims at using a wide range of heterogeneous image understanding datasets with incompatible annotation formats and conflicting label spaces (1). Our methodology derives a unified label space (2) and consolidates supervision (3) so they can be used to simultaneously train an FCN (4) on all datasets. The trained network (5) has better performance on the training datasets, generalizes better to unseen images, and recognizes finer-grained semantic classes.\vspace{-10px}}
	\label{fig:motivation-b}
\end{figure*}
\subsection{Multi-dataset semantic segmentation}
\label{subsec:cross-dataset-ss}
The majority of previous works focus on using multiple datasets with possibly different label spaces, but a single type of supervision,~\ie pixel-level labels. Most of the works solve the challenges that arise from conflicts in label semantics through dataset-based solutions~\cite{ros2016training,lambert2020mseg}, architecture-based solutions~\cite{liang2018dynamic,meletis2018heterogeneous,kalluri2019universal,leonardi2019training,fang2020multi,sun2020real}, or loss-based solutions~\cite{kong2019training,meletis2018heterogeneous,fang2020multi}. In these works, all label spaces of the employed datasets are combined into a common taxonomy, by merging, splitting, or ignoring semantic concepts or by manual re-labeling when needed. Early works extend the conventional FCN architecture with multiple heads/decoders up to one for each dataset~\cite{leonardi2019training}, or multiple (hierarchical) classifiers~\cite{meletis2018heterogeneous}, thereby effectively approaching the problem from the multi-task learning perspective. The authors of \cite{ros2016training} combine six datasets, while in \cite{lambert2020mseg} thirteen datasets are combined to create a large-scale training and testing platform.

Contrary to existing works, the proposed Heterogeneous Training of Semantic Segmentation (HTSS) framework does not require any dataset relabeling and does not ignore classes for simultaneous training of an FCN with multiple datasets. Moreover, it maintains the network architecture, since it solves all label space conflicts at the stage of loss calculation, and therefore the method can be applied to any semantic segmentation network without requiring architectural changes.

\subsection{Semantic Segmentation with weak supervision}
Semantic segmentation is by definition a pixel-based task and it is conventionally realized by training a CNN with per-pixel homogeneous supervision. Previous works have used a diverse set of less detailed (weak) heterogeneous supervision, either to accommodate strong supervision, or independently in a semi/weakly-supervised setting~\cite{zou2020pseudoseg,ibrahim2020semi,meletis2019boosting,papandreou2015weakly,kalluri2019universal}. Several methods generate candidate masks from bounding-box supervision~\cite{zhu2014learning,meletis2019boosting,dai2015boxsup,ibrahim2020semi,meletis2018heterogeneous} using external modules, internal network predictions, or heuristics to refine weak annotations. These masks are used to train networks alone or together with strong supervision. Even weaker forms of supervision have been employed and examples of this include point-level \cite{bearman2016s} and image-level \cite{wang2020weakly,pathak2015constrained,meng2019weakly} annotations, mainly within a multiple instance learning formulation. Finally, methods that use a combination of multiple weaker types of supervision have been proposed, such as bounding boxes and image-level tags \cite{ye2018learning,meletis2019data,papandreou2015weakly,li2019weaklier,li2018weakly}.

Inspired by earlier works, the proposed framework achieves pixel-accurate training using weak supervision by a pre-processing step that generates pseudo-labels and a refinement process during training. Moreover, unlike previous methods, the HTSS framework treats all types of weakly-labeled datasets uniformly and uses them in combination with strongly-labeled datasets.

\subsection{Other related tasks}
Two related semantic segmentation tasks that encapsulate multiple datasets in their formulation are transfer learning~\cite{weiss2016survey} and domain adaptation~\cite{sun2015survey,zhao2020multisource}. These tasks aim at transferring rich knowledge from a source dataset/domain to a target dataset/domain, where knowledge is scarce or even non-existing. They mainly concentrate on the performance in the target domain, which may be available during training, in some limited form. Recently, variations of these tasks also track performance in the source domain and investigate multiple-source versions of the problems~\cite{he2021multi,piva2021exploiting}. The HTSS formulation considers performance on all employed datasets and during training it does not depend on information from the testing datasets explicitly, as in domain adaptation.

The following four tasks are briefly addressed for their relevance with aspects of HTSS. First, multi-dataset semantic segmentation has been tackled in literature using \textit{multi-task learning} ~\cite{ranftl2020depth,nekrasov2019real,kokkinos2017ubernet,vandenhende2021multi,crawshaw2020multitask}, where a network head/branch is devoted to each dataset independently,~\ie segmentation for each dataset is modeled as a separate ``task''. Second, \textit{continual learning} is relevant to the concept of knowledgeability, since in this task new classes are discovered or added during training or inference~\cite{nakajima2019incremental,klingner2020class} and old data may not be available. Third, \textit{self-training} or \textit{pseudo-label} approaches~\cite{saporta2020esl,sanberg2017free,zoph2020rethinking,zou2020pseudoseg,zhan2018mix} have addressed the absence of labels during training for some datasets/domains. Finally, \textit{learning with partial labels}~\cite{zhang2017disambiguation,cid2012proper} is related to conflicting labels spaces. The partial-label formulation associates a training sample with a set of candidate labels among which at most one is the correct label.

This plethora of research shows that training with multiple and heterogeneous datasets is a desired capability of modern training pipelines.

\subsection{Conventional semantic segmentation}
\label{ssec:sem-seg}
We recap here the formulation for the conventional image semantic segmentation, which we use in the following section to define Heterogeneous Training of Semantic Segmentation. The task of Semantic Segmentation~\cite{guo2017aro,long2015fully,garcia2018survey,cordts2016Cityscapes,minaee2020image} involves the per-pixel classification of an image into a predetermined set of mutually-exclusive semantic classes. A semantic segmentation system has a 2-D image $\mathbf{x}$ as input and uses a given label space $\mathcal{L}^\text{pred}$ of semantic classes. The aim of the system is to predict a 2-D matrix $\mathbf{y}^\text{pred}$, where each element corresponds to an image pixel with a semantic class from $\mathcal{L}^\text{pred}$ assigned to it.

In the conventional supervised learning setting, the task entails a dataset $\mathcal{S} =  (\mathcal{D}, \mathcal{L})$, which consists of $N$ image-label pairs ~$\mathcal{D} =  \{\left(\mathbf{x}_i, \mathbf{y}_i\right),~i = 1, \dots, N\}$ and a label space $\mathcal{L} = \left\{l_j,~j = 0, \dots, L\right\}$ of $L$ semantic classes. Every label $\mathbf{y} \in \mathcal{L}^{H \times W}$ is a 2-D matrix with spatial size $H \times W$ and every position corresponds to a single pixel in the image $\mathbf{x}$. It is common that a semantic class (\eg the $l_0$) corresponds to a special label that denotes unlabeled or void pixels. The semantic classes represent semantic entities of a scene,~\eg vehicle, person, tree, or sky. It is essential that all $l_j$ have unambiguous and mutually-exclusive semantic definitions $def(l_j)$ within a dataset,\eg all possible types of \textit{car}s in a dataset should not resemble any of the \textit{truck}s and vice versa. If this does not hold,~\ie annotations are noisy or concepts overlap between classes, then a classifier trained on this dataset may be ``confused'' and the evaluation is inaccurate.

Although in literature, existing datasets strive to define unambiguous and non-overlapping classes within their label space, in practice some ambiguity exists, mainly due to defining semantic classes using a single-word. In this work, we assume that such noise in labels is negligible.

\section{Heterogeneous Training for Semantic Segmentation}
\label{sec:challenges}
In this section we describe the problem and challenges of Heterogeneous Training of Semantic Segmentation (HTSS), prior to presenting the proposed framework in Sec.~\ref{sec:methodology}. The terminology follows the conventional semantic segmentation task described in Sec.~\ref{ssec:sem-seg}.

In the conventional setting described in Section~\ref{ssec:sem-seg}, the formats of the task definition and the given dataset are in full agreement with each other. In this case, the output label space $\mathcal{L}^{\text{pred}}$ can be set to be the dataset label space $\mathcal{L}^{\text{pred}} \equiv \mathcal{L}$ and the predictions are congruent to the per-pixel annotations $\mathbf{y}^{\text{pred}} \cong \mathbf{y}$. The symmetry between the training data and the task goal is advantageous, however it limits the available information that a system can be exposed to,~\eg only one dataset from the first column of Fig.~\ref{fig:datasets-statistics}. These datasets are limited in size and semantic concepts, which renders them inadequate for large-scale, in-the-wild semantic segmentation.

We consider two generalizations to the traditional semantic segmentation problem and encapsulate them in a unified formulation. The first aims at enriching the output semantic space of the task using multiple heterogeneous label spaces. The second intends to increase the amount of available supervision by generalizing the task input to more types of supervision. We incorporate these generalization within a unified formulation by maintaining the task output identical, while relaxing the requirements for the given datasets. This enables potential inclusion of datasets, which are originally created for other scene understanding tasks,~\eg multiple datasets from all columns of Fig.~\ref{fig:datasets-statistics}, when training a network for semantic segmentation.

Heterogeneous semantic segmentation enables trained networks to aggregate information from diverse datasets and it could demonstrate potential improvements in the following three aspects. First, multi-dataset training increases examples for underrepresented classes and provides diversity in recognizable semantics, which could be advantageous for \textit{performance},~\ie segmentation accuracy on seen (training) datasets. Second, the diversity of employed datasets should bring benefits to \textit{generalization},~\ie segmentation accuracy on unseen (testing) datasets. We evaluate this aspect under the \textit{cross-dataset zero-shot} setting~\cite{lambert2020mseg}. Third, network predictions are more fine-grained and semantically rich by incorporating semantics from multiple datasets. As can be observed from Fig.~\ref{fig:datasets-statistics}, label spaces of pixel-labeled datasets are generally smaller than weakly-labeled datasets,~\ie $\mathcal{L}^\text{pixel} \ll \mathcal{L}^\text{bbox} \ll \mathcal{L}^\text{tag}$. We propose a metric, detailed in Sec.~\ref{ssec:knowledgeability}, to quantify the semantic richness of the predicted classes w.r.t. to the segmentation performance for these classes.

\begin{table}
	\small
	\centering
	\setlength\tabcolsep{4.0pt}
	\begin{tabular}{@{}ll@{}}
		\toprule
		Label type & Definition\\
		\midrule
		Pixel (dense) & $\mathbf{y} \in \mathcal{L}^{H \times W}, ~\left |y_k = l_0 \right| \ll \left| y_k \neq l_0 \right|, ~k \in H \times W$\\
		Pixel (coarse) & $\mathbf{y} \in \mathcal{L}^{H \times W}, ~\left |y_k = l_0 \right| \gg \left| y_k \neq l_0 \right|, ~k \in H \times W$\\
		Bound. boxes & $\mathbf{y} = \left\{ \left( l_k, ~\text{bbox-coords}_k \right), ~k = 1, \dots, B \right\}$\\
		Image tags & $\mathbf{y} = \left\{ l_k, ~k = 1, \dots, T \right\}$\\
		\bottomrule
	\end{tabular}
	\caption{A variety of annotation formats are considered. The dataset indexing superscript $(i)$ and the dataset sample subscript $j$ are omitted for clarity. $l_0$ is the \textit{void} class.\vspace{-10pt}}
	\label{tab:annot-formats}
\end{table}

\subsection{Problem formulation}
\label{ssec:prob-form}
Similarly to the conventional segmentation task definition, heterogeneous semantic segmentation aims at predicting a \mbox{2-D} matrix $\mathbf{y}^\text{pred}$ with semantic classes, given a 2-D image $\mathbf{x}$ and a label space $\mathcal{L}^\text{pred}$. Contrary to the traditional single-dataset formulation, we assume that  a collection $\mathbb{S}$ of $D$ heterogeneous datasets is available, where each dataset $\mathcal{S}^{(i)}$ includes $N$ image-label pairs $\mathcal{D}^{(i)}$ and the corresponding label space $\mathcal{L}^{(i)}$ with $L^{(i)}$ semantic classes:
\begin{flalign}
	& \mathbb{S} = \left\{ \mathcal{S}^{(i)}, ~i = 1, \dots, D \right\} ~,\\
	& \mathcal{S}^{(i)} = \left( \mathcal{D}^{(i)}, \mathcal{L}^{(i)} \right) ~,\\
	& \mathcal{D}^{(i)} =  \left\{ \left( \mathbf{x}^{(i)}_j, \mathbf{y}^{(i)}_j \right),~j = 1, \dots, N^{(i)} \right\} ~, \label{eq:dataset-def}\\
	& \mathcal{L}^{(i)} = \left\{ l^{(i)}_m, ~m = 0, \dots, L^{(i)} \right\} ~.
\end{flalign}
%
The type of labels $\mathbf{y}$ considered in this work are provided in Tab.~\ref{tab:annot-formats}.

The goal is to train a system for semantic segmentation, such that it utilizes information from all heterogeneous datasets in $\mathbb{S}$. The system should have a consistent label space and recognize the semantic concepts from all considered label spaces $\mathcal{L}^\text{pred} = \uplus_{i=1}^D \mathcal{L}^{(i)}$. Within the researched problem formulation, the following conditions should hold for the employed datasets.

\noindent\textbf{1. Intra-dataset label space consistency}. Each label space $\mathcal{L}^{(i)}$ should include consistent and mutually-exclusive semantic classes, as explained in Sec.~\ref{ssec:sem-seg}:
\begin{equation}
def\left( l^{(i)}_m \right) \cap def\left( l^{(i)}_n \right) = \emptyset, ~\forall ~i = 1, \dots, D ~.
\label{eq:hypothesis-1}
\end{equation}
where $def(l)$ denotes the proper definition of the semantic class $l$.

\noindent\textbf{2. Condition for weakly-labeled classes}. Any semantic class from a weakly-labeled dataset $\mathcal{S}^{(W)}$ should either be identical to, or contain partially semantics from, a class in a strongly-labeled dataset $\mathcal{S}^{(S)}$:\vspace{-5pt}
\begin{equation}
\exists ~l^{(S)} ~\text{so that} ~def\left( l^{(W)} \right) \subseteq def\left( l^{(S)} \right) ~.
\label{eq:cond-2}
\end{equation}

Note that: i) cond. 2 implies that there must be at least one pixel-labeled dataset available for training, and ii) cond. 1 does not imply inter-dataset label space consistency, which is one of the challenges addressed by our HTSS framework.

\begin{table}
	\footnotesize
	\setlength\tabcolsep{3.5pt}
	\centering
	\begin{tabular}{@{}cc|ll@{}}
		\toprule
		\multicolumn{2}{c|}{Challenges} & \multicolumn{2}{c}{HTSS components}\\
		\midrule
		\makecell[c]{Label\\spaces} & \makecell[c]{Label\\type} & \makecell[l]{Preparation\\(once)} & \makecell[l]{Supervision during\\training (each step)}\\
		\midrule
		\multirow{2}{*}{\rotnighty{\makecell{no\\conflicts}}} & strong & - & standard cross-entropy (CE)\\
		\cmidrule{2-4}
		& \makecell{weak\\(Sec.~\ref{sec:weak-superv})} & \makecell[l]{create pseudo-labels\\(Sec.~\ref{ssec:unify}, Fig.~\ref{fig:temp-convert-supervision})} & \makecell[l]{conditional CE, refine pseu-\\do-labels (Sec.~\ref{ssec:hier-loss}, Fig.~\ref{fig:temp-loss-comps})}\\
		\midrule
		\multirow{2}{*}{\rotnighty{\makecell{with\\conflicts}}} & \makecell{strong\\(Sec.~\ref{sec:label-taxonomy})} & \makecell[l]{build taxonomy\\(Sec.~\ref{ssec:gen-taxonomy}, Fig.~\ref{fig:hierarchy})} & \makecell[l]{supervise semantic atoms\\(Sec.~\ref{ssec:converters}, Fig.~\ref{fig:temp-hierarchical-classifier})}\\
		\cmidrule{2-4}
		& \makecell{weak\\(Sec.~\ref{ssec:general-case})} & \makecell[l]{\textbf{all above}} & \makecell[l]{\textbf{all above}}\\
		\bottomrule
	\end{tabular}
	\caption{Overview of the HTSS components developed in this work. Each row includes methods for combining a strongly-labeled dataset and any other dataset(s) with the type of supervision denoted by the second column.\vspace{-5pt}}
	\label{tab:overview-methods}
\end{table}

\subsection{Challenges}
\label{ssec:challenges}
Heterogeneous semantic segmentation brings forward incompatibilities in the annotation formats and the conflicting label spaces among datasets. The following two paragraphs analyze the related challenges.

\noindent\textbf{Label space conflicts}.
\label{sec:semantic-level-of-detail}
Datasets are annotated over different label spaces on a vast spectrum of semantic detail, as they are collected to serve different purposes, which leads to conflicting or overlapping definitions of classes between datasets. If the class definitions for all labels is matching between datasets then a simple union of the label spaces is feasible. However, this is usually not doable because of potential conflicts. The main source of conflicts stems from partial overlapping semantic class definitions between two arbitrary datasets $\mathcal{S}^{(X)}$ and $\mathcal{S}^{(Y)}$, which is specified by:\vspace{-5pt}

\begin{equation}
	def\left(l^{(X)}\right) \cap def\left(l^{(Y)}\right) \neq \emptyset ~.
\end{equation}

Since, the class definitions can overlap only partially, merging them or including them both in the combined label space will introduce ambiguity to the output label space of a trained network. A special common case occurs when conflicts arise from differences in the semantic level-of-detail between classes. For example, a class $l^{(X)}$ from dataset $\mathcal{S}^{(X)}$ describes a high-level concept which contains many more fine-grained classes for dataset $\mathcal{S}^{(Y)}$, giving:
\begin{equation}
	def\left(l^{(X)}\right) = \bigcup_m def\left(l_m^{(Y)}\right) ~.
\end{equation}
The inclusion of all classes $l^{(X)}$, $l^{(Y)}_m, \forall m$ in a single label space would also imply introducing conflicts.

\noindent\textbf{Annotation format incompatibilities}.
\label{sec:annotation-types}
A plethora of diverse datasets for scene understanding cannot be used in semantic segmentation when using standard training schemes, due to their incompatible annotation formats. The most common of them are shown in Tab.~\ref{tab:annot-formats}. Semantic segmentation is by definition a pixel-wise task, thus it is convenient that training datasets to provide annotations at the same pixel-level format. The spatial localizability of labels from other datasets of Fig.~\ref{fig:datasets-statistics} is not adequate to train a network, because they introduce spatial localization uncertainties during pixel-wise training for segmentation. The incompatibilities in annotation formats are even more pronounced in a multi-dataset training scenario, where a variety of incompatible annotation formats can exist. Heterogeneous semantic segmentation thus requires to extract useful supervision at the pixel level from a much coarser source of information.

\section{Methodology}
\label{sec:methodology}

The development of our methodology for heterogeneous multi-dataset training abides to the design principle of maintaining the established single-backbone FCN pipeline. This desideratum enables straightforward applicability of the proposed methods to current or future FCN-based architectures, and scalability to an arbitrary number of datasets. 

%
\begin{figure}
	\centering
	\includegraphics[width=0.89\linewidth]{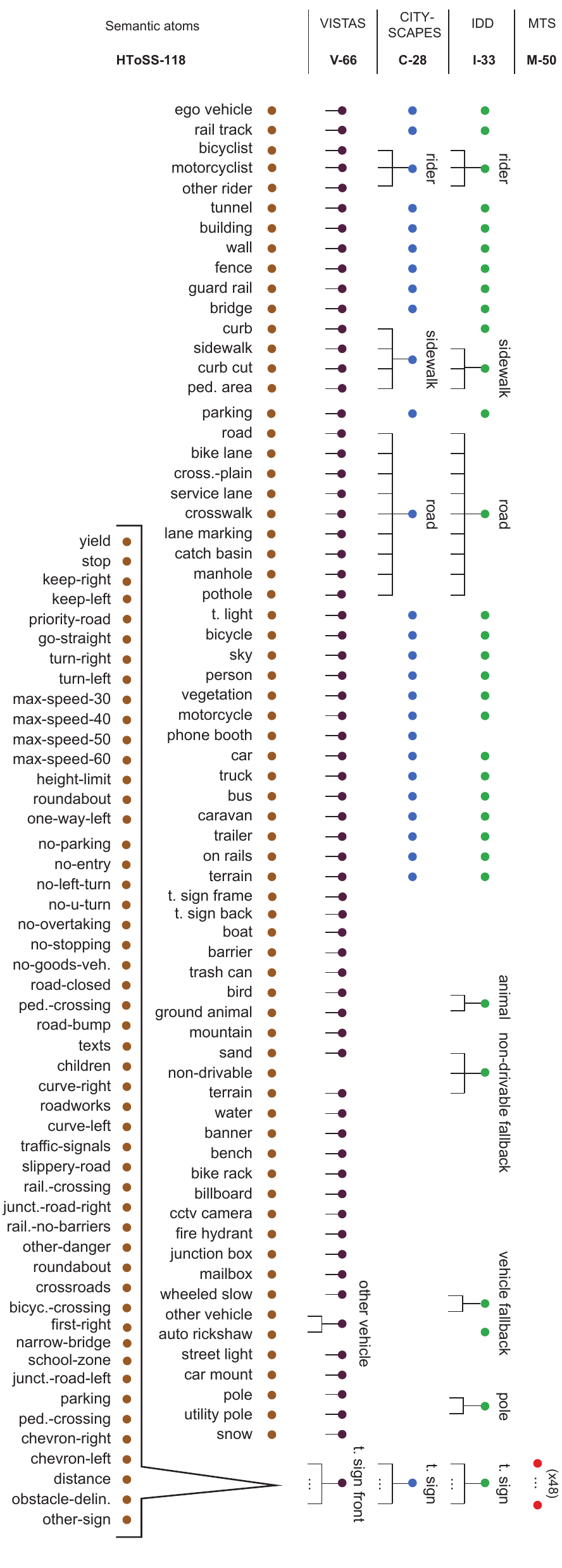}
	\caption{Combined taxonomy of 118 \textit{semantic atoms} (\textit{void} atom is not shown) merging a total of 174 semantic classes from Cityscapes (27), Vistas (65), IDD (33), and MTS (50) datasets. Each dataset section corresponds to the grouping of the combined taxonomy labels in order to apply strong supervision from the respective dataset.}
	\label{fig:hierarchy}
\end{figure}
\begin{figure*}
	\centering
	\includegraphics[width=0.95\linewidth]{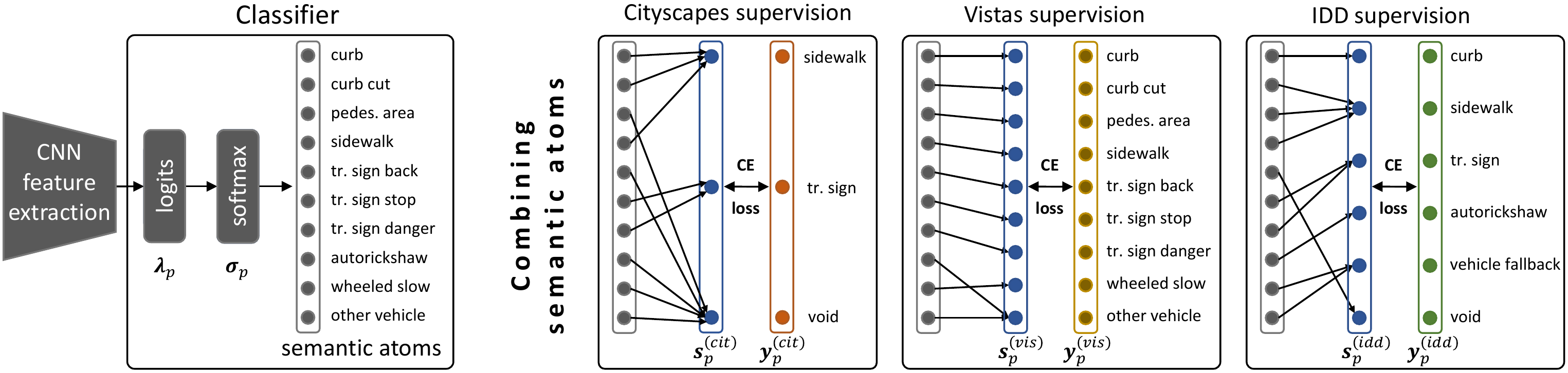}
	\caption{The HTSS classifier is supervised using the standard cross-entropy (CE) loss between the \textit{semantic atoms} vector and a different ground truth vector per dataset (in boxes). This is achieved by combining the CNN output onto a new multinomial distribution vector that matches each dataset's labels. A selection of classes from Fig.~\ref{fig:hierarchy} are shown in order to explain the procedure of probabilities accumulation.\vspace{-10pt}}
	\label{fig:temp-hierarchical-classifier}
\end{figure*}

The proposed Heterogeneous Training of Semantic Segmentation (HTSS) framework addresses the challenges of Sec.~\ref{ssec:challenges} by introducing a methodology for combing disjoint or conflicting label spaces (Sec.~\ref{sec:label-taxonomy}), training a single-backbone network with strong and weak supervision simultaneously (Sec.~\ref{sec:weak-superv}, Sec.~\ref{ssec:general-case}). An overview of the methodology is depicted in Fig.~\ref{fig:motivation-b}. Tab.~\ref{tab:overview-methods} associates the components of our methodology with the challenges they address.

\subsection{Combine datasets with different label spaces}
\label{sec:label-taxonomy}
This subsection describes our approach for training a single-backbone, single-classifier FCN on multiple pixel-labeled datasets, which have disjoint or conflicting semantic label spaces. The approach enables training on an arbitrary number of datasets and it consists of two steps.

\noindent \textbf{Before training}.
First, the label spaces of the training datasets have to be consolidated into a unified taxonomy (Sec.~\ref{ssec:gen-taxonomy}). This procedure can be automated if the label spaces are part of a semantic ontology or hierarchical lexical database (\eg WordNet~\cite{miller1995wordnet} or BabelNet~\cite{navigli2010babelnet}). For example, the classes of ImageNet~\cite{deng2009imagenet} are organized according to the WordNet hierarchy. An ontology~\cite{kulmanov2020semantic} contains lexico-semantic relations (\eg hypernymy/hyponymy, holonymy/meronymy) and the semantic relatedness (\mbox{is-a}, \mbox{has-a}) of concepts, which can be used to group classes together or split them into finer semantic concepts. In this work, the label spaces of the employed datasets are not constructed according to an ontology, thus we used lexico-semantic relations from \textit{WordNet}, \textit{BabelNet}, \textit{thesaurus.com}, and \textit{merriam-webster.com} to generate a unified taxonomy of semantic concepts and minimize manual effort.

\noindent \textbf{During training}.
A single classifier makes predictions over the consistent, unified taxonomy of labels obtained from the previous step. Only during training, these predictions are mapped back to the label space of each dataset with specific per-dataset converters (Sec.~\ref{ssec:converters}), in which case the segmentation loss can be directly applied.

\subsubsection{Generate unified taxonomy of label spaces}
\label{ssec:gen-taxonomy}
When combining multiple datasets it is highly probable that classes between datasets have conflicting definitions, as described in Sec.~\ref{ssec:challenges}.
In order to solve these conflicts and generate a unified taxonomy, we introduce the concept of the \textit{semantic atoms}. A \textit{semantic atom} $\alpha$ is a fine-level semantic primitive (class), whose definition coincide either fully or partially to a definition of a semantic class from a dataset in $\mathbb{S}$. A set of properly chosen \textit{semantic atoms} $\mathcal{A} = \left\{ \alpha_m, ~m = 1, \dots, A \right\}$ fully covers the semantics of all employed datasets without any conflicts. Using the ontologies from the aforementioned online sources we automate the construction of set $\mathcal{A}$. The set is subsequently validated by a human merely for inconsistencies due to ambiguities in their (natural language) description (Algorithm~\ref{alg:sematoms}).

\begin{algorithm}[t]
\DontPrintSemicolon
\KwData{label spaces from all datasets $\mathcal{L}^{(i)}$, $i = 1, \dots, D$}
\KwResult{the set of \textit{semantic atoms} $\mathcal{A}$}
\tcc{\scriptsize Initialize semantic atoms with the multi-set of all labels, then remove ones not complying to conditions.}
all\_labels $\leftarrow$  $\uplus_{i=1}^D \mathcal{L}^{(i)}$\;
\Repeat{no label is removed from all\_labels}{
	\tcc{\mbox{\scriptsize generate combinations of labels pairs}}
	pairs = generate\_combinations(all\_labels)\;
	\For{pair in pairs}{
		ls, lo = pair\;
		\If{(synonym(ls, lo) or hypernym(ls, lo) or holonym(ls, lo))}{
			all\_labels.remove(ls)\;
			break\;
		}
	}
}
$\mathcal{A}$ $\leftarrow$  all\_labels\;
\caption{\label{alg:sematoms} Combine the labels from existing datasets to create the set of \textit{semantic atoms} using the lexico-semantic relations from \textit{WordNet}, \textit{BabelNet}, \textit{thesaurus.com}, and \textit{merriam-webster.com}.}
\end{algorithm}

The following three properties hold for all \textit{semantic atoms}. First, each \textit{semantic atom} should have a concise and unique semantic definition that does not overlap with any of the other \textit{semantic atoms}:
\begin{equation}
\text{def}\left(\alpha_k\right) \cap \text{def}\left(\alpha_m\right) = \emptyset, ~\forall ~k \neq m ~.
\end{equation}
Second, its definition matches fully or partially to a definition of a semantic class from a dataset and thus every \textit{semantic atom} corresponds to (\textit{is-a}, \textit{hyponym}, \textit{meronym}) at most one semantic class:
%
\begin{equation}
\text{def}(\alpha_m) \subseteq \text{def}(l_n), ~\forall \alpha_m \in \mathcal{A}, ~l_n \in \mathfrak{L} \label{eq:sem-atom-def} ~,
\end{equation}
where $\mathfrak{L} = \cup_i \mathcal{L}^{(i)}, ~i = 1, \dots, D$ is the set of labels from all label spaces of the datasets to be combined.
Third, the set of all \textit{semantic atoms} should completely describe the semantics of all datasets, which yields that every semantic class $l_n$ consists of (\textit{has-a}, \textit{hypernym}, \textit{holonym}) at least one \textit{semantic atom}:
%
\begin{equation}
\text{def}(l_n) = \bigcup_{m \in M_n} \text{def}(\alpha_m) , ~\forall l_n \in \mathfrak{L} ~,
\end{equation}
where $M_n$ is a set of indices corresponding to class $l_n$.

As long as the manual process for the extraction of \textit{semantic atoms} is completed, the taxonomy of semantic classes from all datasets can be generated (see Figure~\ref{fig:hierarchy}). Then, we can train a single-classifier FCN using the \textit{semantic atoms} as output classes and supervise this CNN using the original label spaces from each dataset, by combing the \textit{atoms} using the generated unified taxonomy as described in the next section.

\subsubsection{Supervise \textit{semantic atoms} with the original label spaces}
\label{ssec:converters}
Having extracted the set of \textit{semantic atoms} $\mathcal{A}$ that fully covers the semantics of the employed datasets $\mathbb{S}$, we can train a single-backbone, single-classifier FCN with output label space $\mathcal{A}$. This procedure is shown in Fig.~\ref{fig:temp-hierarchical-classifier} for a selection of \textit{semantic atoms} from Fig.~\ref{fig:hierarchy}. The output of the classifier for spatial position (pixel) $p$ and dataset $i$ is the categorical probability vector $\bsigma^{(i)}_p \in \left[0, 1\right]^A$, of cardinality $A = \mathcal{A}$, where each element corresponds to the probability of a \textit{semantic atom} in $\mathcal{A}$. Since $\bsigma^{(i)}_p$ represents a categorical probability it holds that $\sum_m \sigma^{(i)}_{p, m}=1$. In the following, we describe how $\bsigma^{(i)}_p$ is transformed to be compatible with the original label space $\mathcal{L}^{(i)}$ of each dataset of the taxonomy, in order to train the classifier using the conventional cross-entropy loss.

Conceptually, for each supervising dataset $i$, we are going to map the categorical output $\bsigma_p^{(i)}$ to the categorical labels. Via this mapping the labels of the original dataset can supervise (in)directly the training of the semantic atoms. The extraction of \textit{semantic atoms} induces a collection of sets $\{G^{(i)}_m, ~i = 1, \dots, D, ~m = 0, \dots, L^{(i)}\}$. Each $G^{(i)}_m$ contains the \textit{semantic atoms} that correspond to class $l^{(i)}_m$ from dataset $i$. According to the taxonomy construction process (Section~\ref{ssec:gen-taxonomy}), the extracted \textit{semantic atoms} fully describe the semantics of all classes from all selected datasets. As a consequence, an arbitrary dataset class is represented by either a single or a combination of \textit{semantic atom}(\textit{s}). Using this property, we can partition $\bsigma^{(i)}_p$ into groups according to sets $G^{(i)}_m$ and accumulate their probabilities into a reduced vector $\bm{s}_p^{(i)} \in {[0, 1]}^{L^{(i)}}$ for each dataset $i$. This process can be written concisely as:
\begin{equation}
s^{(i)}_{p, m}(\bsigma_p) = \sum_{\alpha \in G^{(i)}_m} \sigma^{(i)}_{p, \alpha} ~,
\end{equation}
where for now, an integer number is assigned as ``name'' to every \textit{atom} $\alpha$, so that they can be used for indexing $\mathcal{A} = \{1, 2, \dots, A\}$. Since $\bsigma^{(i)}_p$ is a categorical distribution, then $\bm{s}^{(i)}_p$ is also a categorical distribution. Moreover, it contains classes that correspond one-by-one to the ground truth $\bm{y}^{(i)}_p$. Thus, they can be used in the standard cross-entropy loss formulation.

During batch-wise training, a batch can contain images from many datasets. Without loss of generality, we will formulate the cross-entropy loss for a single image $j$ from dataset $i$ in the batch. The label $\bm{y}^{(i)}_j$ (Eq.~\eqref{eq:dataset-def}) has shape $H \times W \times L^{(i)}$ (one-hot encoding), and the output $\bsigma^{(i)}_j$ has shape $H \times W \times A$. By using a single index $p \in P$ to enumerate spatial positions ($H, W$) and omitting from the notation of $\bm{y}$ and $\bsigma$ the dataset index $i$ and image index $j$, the cross-entropy loss can be expressed as:
\begin{equation}
\text{Loss}_j \left(\bm{y}, \bsigma \right) = -\dfrac{1}{\left|P\right|} \sum_{p \in \mathcal{P}} \sum_m y_{p, m} \log s^{(i)}_{p, m} ~. \label{eq:loss-per-image}
\end{equation}

In the following, we derive the gradients of the loss wrt. the logits of the network and prove that our method is a generalization of the standard formulation. As this is independent of the dataset and the position indices, we drop them for minimizing notation clutter. The logits $\bm{\lambda} \in \mathbb{R}^A$ are the input of the softmax $\bsigma$, where $\sigma_i (\bm{\lambda}) = e^{\lambda_i} / \sum_j e^{\lambda_j}$ and the converted outputs of the network can be expressed as $ \bm{s}\left(\bsigma\left(\bm{\lambda}\right)\right)$. Using the backpropagation rule the gradient of the loss wrt. the logits is:
\begin{equation}
\frac{\partial \text{Loss}}{\partial \bm{\lambda}} = \frac{\partial \text{Loss}}{\partial \bm{s}} \cdot \frac{\partial \bm{s}}{\partial \bsigma} \cdot \frac{\partial \bsigma}{\partial \bm{\lambda}} ~. \label{eq:part-der}
\end{equation}
Since each pixel in the annotations has a single class (one-hot),~\ie $y_m = 1$ for class $m = m^*$ and $y_m = 0,~m \ne m^* $, the loss of Eq.~\eqref{eq:loss-per-image} (omitting the summation over positions $p$) reduces to $-\sum_{m} \llbracket m = m^* \rrbracket \log s_m = -\log s_{m^*}$, where $\llbracket \cdot \rrbracket$ is the Iverson bracket. It is easy to show that the partial derivatives of the factors in Eq.~\eqref{eq:part-der} are:
$\partial \text{Loss}/\partial s_i = -\llbracket i = i^* \rrbracket/\sigma_{i} $,
$\partial s_i / \partial \sigma_j = \llbracket j \in G_i \rrbracket$, and 
$\partial \sigma_i / \partial \lambda_j = \sigma_i \left( \llbracket i = j \rrbracket - \sigma_j \right)$. Substituting these into Eq.~\eqref{eq:part-der} it yields:
\begin{equation}
\frac{\partial \text{Loss}}{\partial \lambda_m} = \sigma_m - \llbracket m \in G_{m^*} \rrbracket ~,
\end{equation}
which is a mere generalization of the loss derivative in the original FCN framework, which is $\partial \text{Loss} / \partial \lambda_m = \sigma_m - \llbracket m = m^* \rrbracket $. This property ensures comparable gradient flows between FCN and our framework, and thus no architectural changes or loss modifications are needed.
\begin{figure}
	\centering
	\includegraphics[width=1.0\linewidth]{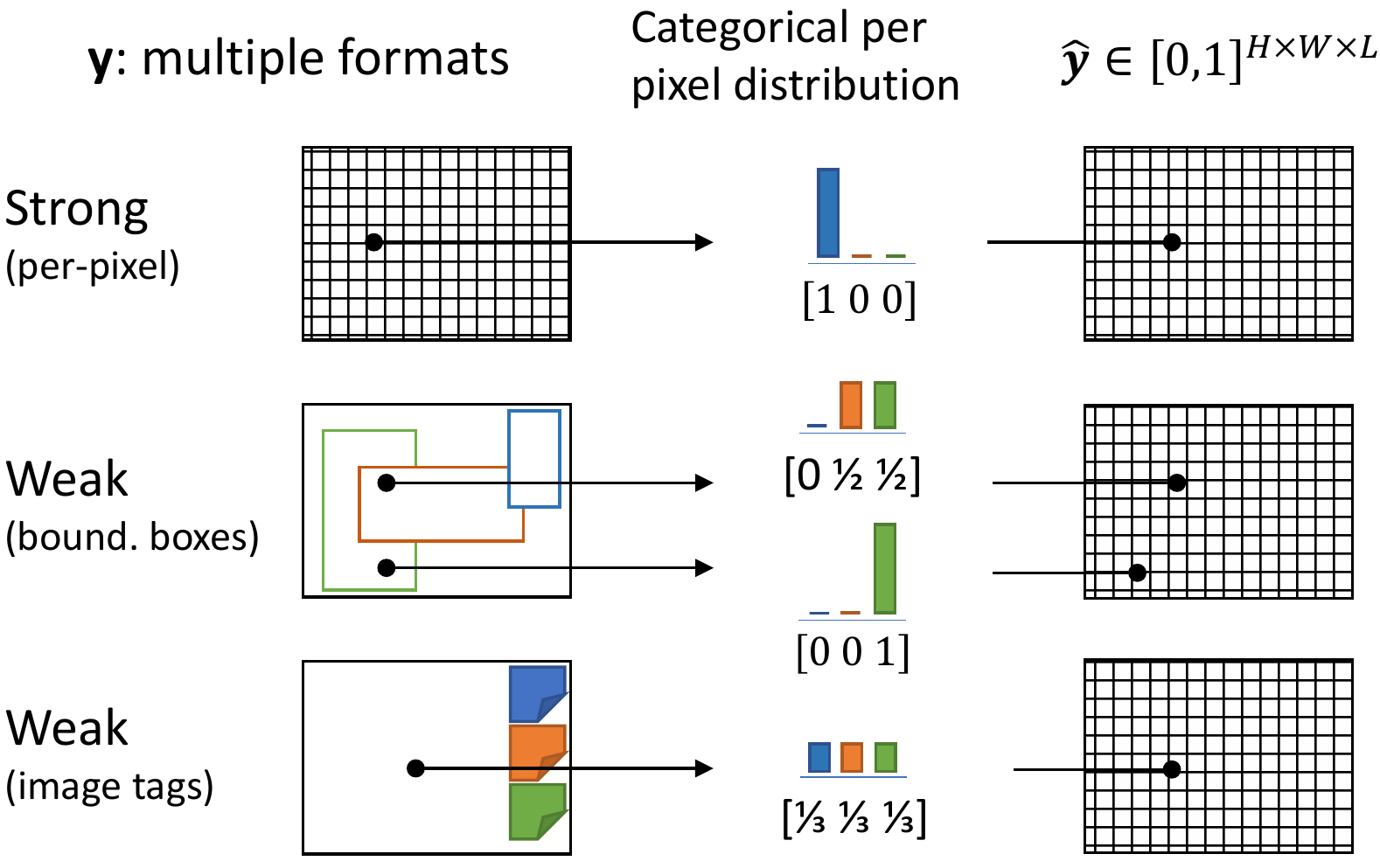}
	\caption{Creation of pseudo labels from weakly-annotated datasets before training. For each pixel in the image a categorical probability vector is created with elements corresponding to the set of annotated classes ($L = 3$ in this example). For each position of $\hat{\mathbf{y}}$, each element along the third dimension is assigned a probability,~\ie the normalized number of respective bounding boxes covering that pixel. During training, these pseudo labels are further refined using input from the classifier itself, see Fig.~\ref{fig:temp-loss-comps}.}
	\label{fig:temp-convert-supervision}
\end{figure}

\subsection{Combine datasets with different annotation types}
\label{sec:weak-superv}
This section describes how a single-backbone, single-classifier FCN is trained on multiple weakly- or strongly- labeled datasets. For now, we assume that the label spaces of all datasets are identical. This limitation is lifted in Sec.~\ref{ssec:general-case}, where combined training with different annotation types and conflicting labels spaces is investigated.

As explained in Sec.~\ref{ssec:challenges}, the spatial localization of annotations in weakly-labeled datasets,~\eg bounding boxes and image tags, is inadequate for providing useful pixel-level supervision. However, if properly conditioned or refined, these spatial locations have the potential to provide helpful cues for increasing segmentation performance. A two-step approach is followed, abiding to the FCN design principles without adding extra modules to the network. First, as described in Sec.~\ref{ssec:unify}, weak annotations from all datasets are converted to pseudo per-pixel labels, so they can be seamlessly used together with pixel labels from strongly-labeled datasets for pixel-wise training. Second, during each training step, the pseudo labels are refined, using only information from the network from this step, without requiring any external knowledge. The process is analyzed in Sec.~\ref{ssec:hier-loss}.

\subsubsection{Unifying weak and strong annotations}
\label{ssec:unify}
The objective of unifying heterogeneous annotations is to transform the weak annotations into per-pixel pseudo labels, so they can be integrated in the pixel-wise training loss. The pseudo labels are then refined during training, to provide a best-effort approximation of the ideal fine labels (Sec.~\ref{ssec:hier-loss}).

We consider weak supervision from bounding boxes and image-level tags, as listed in Tab.~\ref{tab:annot-formats}. Bounding box annotations have orthogonal boundaries, which rarely match the smooth object boundaries,~\eg poles, humans, bicycles, while image tags have even coarser localization. We treat image tags as bounding boxes that extend to the whole image. This forms a basis to handle both annotation formats within a common formulation.

The core of the method involves representing the per-pixel label as a categorical probability vector $\hat{\bm{y}}_p \in \left[0, 1\right]^L$ over the set of all classes $\mathcal{L}$ of the dataset it belongs to. This choice enables including information from all bounding boxes, even if they heavily overlap, and does not require hard choices to assign a single class to each pixel,~\eg assigning randomly or by heuristics. The algorithm is described in the following paragraph and visualized in Figure~\ref{fig:temp-convert-supervision}.

Given a weak label $\mathbf{y}$, we initiate a 3-D label canvas, with two spatial dimensions, being equal to the label size, and a depth dimension with size $L$ for the semantic classes. Each bounding box in $\mathcal{B}$ casts a unity vote to all spatial locations (pixels) being covered by it, at a single position in the depth dimension that corresponds to the semantic class of the bounding box. After the voting is completed from all boxes, the label canvas is normalized along the depth dimensions (semantic classes) to unity, by dividing by the sum of votes for that position. Then, another 2-D slice is concatenated along the same spatial dimensions, corresponding to the \textit{unlabeled} semantic class. Finally, for pixels that are not covered by any bounding box, the \textit{unlabeled} probability is set to unity. At this point, a valid categorical probability vector for each image pixel is obtained, which can be directly used as-is in the conventional per-pixel cross-entropy loss.

\subsubsection{Supervising \textit{semantic atoms} with unified annotations}
\label{ssec:hier-loss}
In the previous section, it was sketched how weak annotations are transformed into per-pixel pseudo-labels that are used in the cross-entropy loss formulation after their spatial locality is refined. Here, the refinement of these coarse labels is described. The refinement is performed in an online fashion during training and generates more (spatially) localized labels for supervision. It is achieved by applying two conditions to pseudo-labels that omit supervision for uncertain or ambiguous pixels,~\eg pixels that may reside outside of an object.

\begin{figure}
	\centering
	\includegraphics[width=1.0\linewidth]{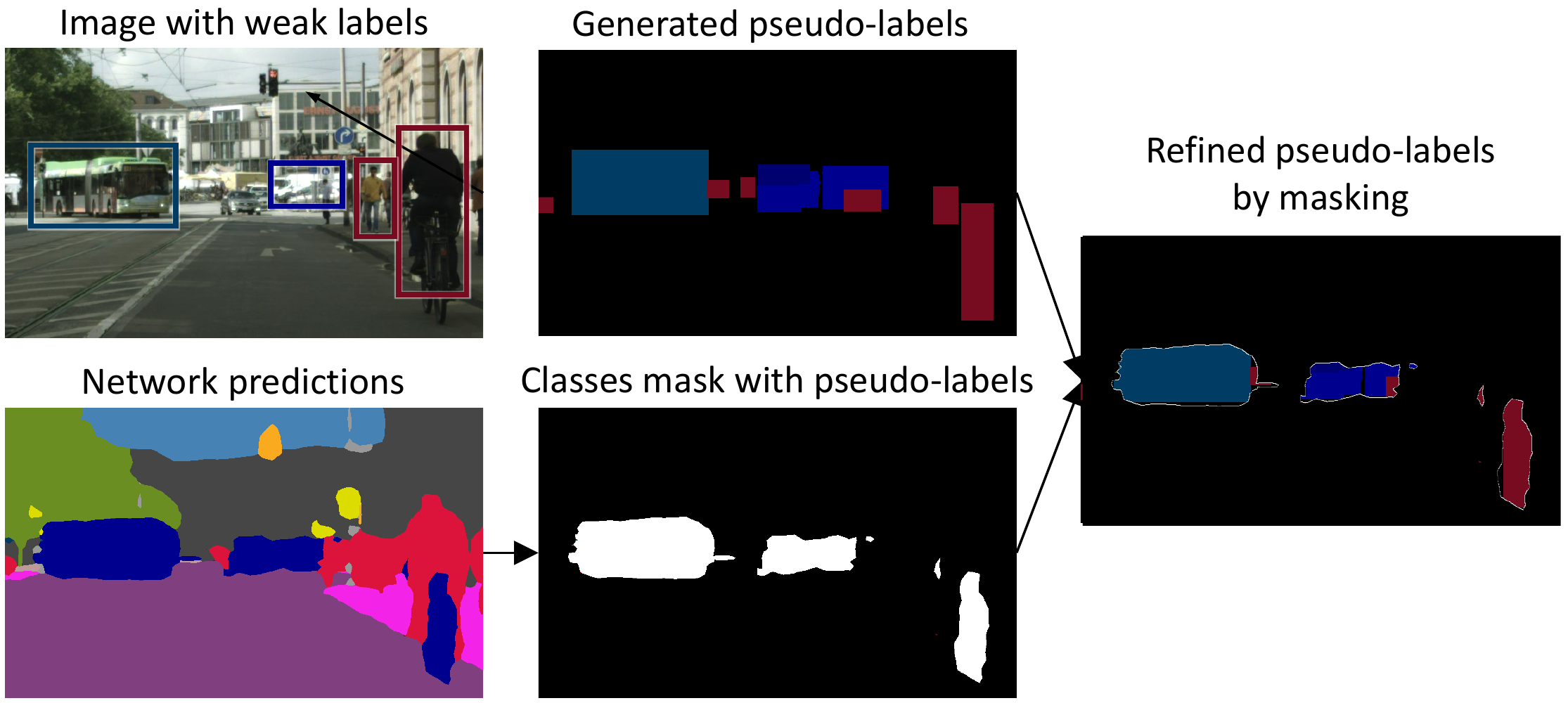}
	\caption{Refinement of pseudo labels during each training step using the predictions of the network in that step.\vspace{-12pt}}
	\label{fig:temp-loss-comps}
\end{figure}

We assume that a collection of $\mathbb{S}^{(S)}$ strongly-labeled and $\mathbb{S}^{(W)}$ weakly-labeled datasets are given and all datasets have an identical label space. First, the weak labels $\bm{y}^{(W)}$ of $\mathbb{S}^{(W)}$ are converted into per-pixel pseudo-labels $\hat{\mathbf{y}}$, using the procedure of Sec.~\ref{ssec:unify}. Then, during a training step, we rely on the network predictions, which is the best estimate it can be attained, to improve the pseudo-labels. If $\bsigma_p \in [0, 1]^L$ is the softmax output for position $p$ for an image with weak labels, the predictions can be expressed as $\pi_p = \argmax_m \sigma_{p, m}$. The refined pseudo-labels $\tilde{\mathbf{y}}_p$ are obtained for pixel $p$ by keeping the pseudo-label if the prediction agrees and the corresponding probability is higher than a threshold $T$, as follows:
\begin{equation}
\tilde{\mathbf{y}}_p = 
\begin{cases}
\hat{\mathbf{y}}_p, & \text{if} ~\pi_p = \argmax_m \hat{y}_{p, m} ~\text{and} ~\sigma_{p, \pi_p} \ge T,\\
\text{unlabeled}, & \text{otherwise}
\end{cases}
\label{eq:conds}
\end{equation}
The first condition of Eq.~\eqref{eq:conds} is illustrated in Fig.~\ref{fig:temp-loss-comps}. The second condition refers to the magnitude of the probability of the predictions, which can be view as a measure of confidence. Specifically, a heuristically chosen threshold $T$ is used, which should be exceeded by the probability of the highest predicted class, in order to be deemed reliable. This threshold provides a good trade-off between utilizing enough weak labels, while maintaining their confidence high. For the experiments, we have empirically chosen  $T = 0.9$.

The final loss for the batch with images from both weakly-labeled  $(W)$ and strongly-labeled $(S)$ datasets is computed as:
\begin{equation}
	\text{Loss} = - \sum_{p \in \mathcal{P}} \sum_j z_{p,j} \log \sigma_{p,j} ~,
	\label{eq:loss-strong-weak}
\end{equation}
where $\mathcal{P} = \mathcal{P}^{(S)} \cup \mathcal{P}^{(W)}$ is the set of all pixels and $z_{p, j}$ is defined as:
\begin{equation}
	\bm{z}_p = 
	\begin{cases}
		\frac{1}{\abs{\mathcal{P}^{(S)}}} \bm{y}^{(S)}_p, & p \in \mathcal{P}^{(S)} \\
		\frac{1}{\abs{\mathcal{P}^{(W)}}} \tilde{\bm{y}}^{(W)}_p, & p \in \mathcal{P}^{(W)} ~.
	\end{cases}
\end{equation}

\subsection{Conflicting label spaces and annotation types}
\label{ssec:general-case}
Sec.~\ref{sec:label-taxonomy} proposed a solution for label-space conflicts considering only strongly-labeled datasets. Sec.~\ref{sec:weak-superv} proposed a solution for training networks with multiple annotation types considering only datasets with identical label spaces. The combination of the two approaches that is able to simultaneously train networks with any datasets (case of last row of Tab.~\ref{tab:overview-methods}) is described in this section.

The extraction of \textit{semantic atoms} (Sec.~\ref{ssec:gen-taxonomy}) and the conversion of weak to per-pixel annotations (Sec.~\ref{ssec:unify}) can be directly applied to the selected datasets. However, for supervising the \textit{semantic atoms}, the formulas of Sec.~\ref{ssec:converters}, and ~\ref{ssec:hier-loss} cannot be directly applied for any \textit{semantic atom} and a small set of them requires a different handling. According to this distinction, the \textit{semantic atoms} are split into two sets $\mathcal{A}^a$ and  $\mathcal{A}^s$ with classes that need special care. Each \textit{atom} in $\mathcal{A}^a$ is either a class with only strong labels, or with weak and strong labels from different datasets. Each \textit{atom} in $\mathcal{A}^s$ is strictly a class with weak labels for which the condition of Eq.~\eqref{eq:cond-2} holds. The localization cues of the \textit{atoms} in $\mathcal{A}^s$ are extremely sparse, due to their weak annotations and the fact that they do not appear in strongly-labeled datasets. Consequently, the refinement step (Eq.~\eqref{eq:conds}) is ineffective for pixel-accurate segmentation. As a solution, for these \textit{atoms} (\eg the traffic sign subclasses in the taxonomy of Fig.~\ref{fig:hierarchy}), we use the parent (strongly-labeled) classes $\mathcal{A}^p$ (\eg \textit{traffic sign front}) as cues for pixel-accurate segmentation. Then, fine-grained semantics can be attained using classification over $\mathcal{A}^s$.

\begin{figure}
	\centering
	\includegraphics[width=1.0\linewidth]{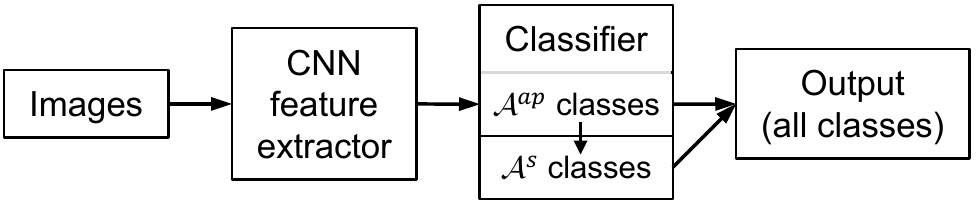}
	\caption{Classifier structure in case of conflicting label spaces and mixed (strong and weak) supervision. The final predictions (right block) are the predictions of classifier part $\mathcal{A}^{ap}$, where the pixels that are assigned classes belonging to classifier part $\mathcal{A}^p$ are replaced by subclass predictions of classifier part $\mathcal{A}^s$.\vspace{-5pt}}
	\label{fig:classifiers-structure}
\end{figure}

The predictions of the classifier, as illustrated in Fig.~\ref{fig:classifiers-structure}, are over two sets of classes: $\mathcal{A}^{ap} = \mathcal{A}^{a} \cup \mathcal{A}^p$ and $\mathcal{A}^s$. For the subclasses in $\mathcal{A}^s$ the relationship between them and the parent classes $\mathcal{A}^p$ is leveraged. Specifically, the predictions of the corresponding parent class in $\mathcal{A}^p$ are used to provide cues for the refinement process (for example the predicted segmentation masks of the \textit{traffic sign front} class are used to refine the bounding boxes of the traffic sign subclasses of Fig.~\ref{fig:hierarchy}). The classifier is trained using the losses from Eq.~\eqref{eq:loss-per-image} and~\eqref{eq:loss-strong-weak}. During inference, the subclasses of $\mathcal{A}^s$ simply replace their parent classes from $\mathcal{A}^p$ in the final predictions.

\section{Experimental evaluation}
\label{sec:experimentation}
We conduct extensive experimentation with various dataset combinations to validate our methodology for multi-dataset simultaneous training with mixed supervision and conflicting label spaces. The results are assessed on three directions: i) segmentation performance on seen datasets,~\ie validation/testing splits of training datasets, ii) generalization on unseen datasets,~\ie splits of datasets not used during training, and iii) semantic knowledgeability: number of recognizable semantic concepts. Sec.~\ref{ssec:metrics} describes the evaluation metrics and Sec.~\ref{ssec:setup} discusses the technical details of our experiments. The following three Sections~\ref{ssec:exps-combine-label-spaces},~\ref{ssec:exps-strong-weak-nonconfl},~\ref{ssec:exps-combine-supervision} investigate the three scenarios in multi-dataset training appearing in Table~\ref{tab:overview-methods}. Finally, Sec.~\ref{ssec:exps-ablations} contains ablations for specific design choices in this work. A selection of diverse datasets for street scene understanding with strong and weak supervision are used. An overview of the employed datasets is shown in Table~\ref{tab:datasets-overview}. For each dataset a collection of label spaces is defined.

\subsection{Evaluation metrics}
\label{ssec:metrics}
We use two metric families to quantify the performance of the models. The first family consists of the standard Intersection over Union (IoU) metric~\cite{lin2014microsoft,cordts2016Cityscapes} and averages of it (arithmetic -- mIoU) to summarize performance and generalizability over multiple classes and across different datasets. The second family is based on a new metric, namely \textit{Knowledgeability} that we define in the following subsection.

\begin{table}
	\small
	\centering
	\setlength\tabcolsep{3.5pt}
	\begin{tabular}{@{}lHcrc@{}H@{}}
		\toprule
		Dataset name & Abv. & Labels & \multicolumn{1}{c}{\# imgs} & \makecell{Train./Eval.\\Lab. space} & \makecell{L. space\\Subsets}\\ 
		\midrule
		\textbf{Training datasets}\\
		~~Cityscapes~\cite{cordts2016Cityscapes} & C & pixel & 2,975 & C-28 & C-20 \\
		~~Cityscapes Coarse~\cite{cordts2016Cityscapes} & CC & \makecell{pixel, bbox} & 19,997 & C-28 & C-20 \\
		~~Cityscapes T. Signs~\cite{meletis2018heterogeneous} & CT & pixel & 2,975 & CT-62 & C-20, C-20\\
		~~Mapillary Vistas~\cite{neuhold2017mapillary} & V & pixel & 18,000 & V-66\\
		~~Indian Driving D.~\cite{varma2019idd} & I & pixel & 20,000 & I-33\\
		~~EuroCity-Persons~\cite{braun2019eurocity} & E & bbox & 40,000 & E-2\\
		~~Map. Traffic Signs~\cite{ertler2020mapillary} & T & bbox & 36,589 & T-51 \\
		~~Open Images~\cite{kuznetsova2020open} & O & \makecell{bbox, im. tag} & 100,000 & O-51\\
		\midrule
		\textbf{Testing datasets}\\
		~~Cityscapes (val)~\cite{cordts2016Cityscapes} & Cv & pixel & 500 & C-20 & CC-10, E-2\\
		~~Cityscapes T. Signs~\cite{meletis2018heterogeneous} & Cv & pixel & 500 & CT-34 & CT-20\\
		~~Mapil. Vistas (val)~\cite{neuhold2017mapillary} & Vv & pixel & 2,000 & V-66 & CC-10, E-2\\
		~~IDD (val)~\cite{varma2019idd} & Iv & pixel & 2,036 & I-33 &  CC-10, E-2 \\
		\midrule
		\multicolumn{3}{l}{\textbf{Generalization (unseen) datasets}}\\
		~~Wild Dash (val)~\cite{zendel2018wilddash} & W & pixel & 70 & W-19 \\
		~~KITTI (train)~\cite{geiger2013vision} & K & pixel & 200 & K-20 \\
		\bottomrule
	\end{tabular}
	\caption{Overview of employed datasets for experimentation. The type of annotations, the number of images and the label spaces are shown. The networks are trained with the label spaces of the training datasets. The evaluation label spaces may be smaller than the training counterparts for the same dataset, due to missing classes in testing/generalization datasets or smaller official splits.} 
	\label{tab:datasets-overview}
\end{table}

\begin{table*}
	\centering
	\small
	\setlength\tabcolsep{3.5pt}
	\begin{tabular}{ccc|c||*{5}l|*{5}l}
		\toprule
		\multicolumn{3}{l|}{Train datasets} & \multirow{3}{*}{\makecell{Output\\label\\space}} & \multicolumn{5}{c|}{Accuracy [mIoU \%]} & \multicolumn{5}{c}{Knowledgeability [$\mathcal{K}^{L(\text{dataset})}$ \%]}\\
		\multirowrot{2}{~~Citys} & \multirowrot{2}{~IDD} & \multirowrot{2}{~~Vistas} &  & \multicolumn{2}{c}{Zero-shot Generalization} & \multicolumn{3}{c|}{Val-split Generalization} & \multicolumn{2}{c}{Unseen datasets} & \multicolumn{3}{c}{Seen datasets}\\
		\cmidrule(l{3pt}r{3pt}){5-6} \cmidrule(l{3pt}r{3pt}){7-9} \cmidrule(l{3pt}r{3pt}){10-11} \cmidrule(l{3pt}r{3pt}){12-14}
		& & & & WildDash & KITTI & Citys & IDD & Vistas & WildDash & KITTI & Citys & IDD & Vistas\\
		\midrule
		\ding{51} & & & C-20 & 27.8 & 48.0 & 63.0 & 29.8 & 22.5 & 39.2 & 43.1 & 50.3 & 30.4 & 26.0\\ 
		& \ding{51} & & I-33 & 36.8 & 40.2 & 47.3 & \underline{63.4} & 23.7 & 48.6 & 50.9 & 55.3 & \underline{65.0} & 49.1\\ 
		& & \ding{51} & V-66 & 42.4 & 49.5 & 67.6 & 41.0 & 40.9 & 53.2 & 57.5 & 60.3 & 53.0 & 63.5\\ 
		\midrule
		\ding{51} & \ding{51} & & HTSS-34 & 36.3 & 55.3 & 73.0 & 61.3 & 26.4 & 49.9 & 50.5 & 56.8 & 58.5 & 49.8\\
		\ding{51} & & \ding{51} & HTSS-66 & \underline{44.4} & 51.6 & 61.2 & 43.7 & \underline{43.1} & 60.0 & 62.4 & 62.3 & 61.0 & 62.9\\
		& \ding{51} & \ding{51} & HTSS-68 & 44.0 & 53.8 & 69.2 & 58.0 & 42.6 & 59.5 & 63.0 & 60.1 & 62.4 & 63.0\\
		\ding{51} & \ding{51} & \ding{51} & HTSS-68 & \underline{44.4} {\tiny \color{ForestGreen} $\left\lceil +16.6 \right\rceil$} & \underline{56.5} {\tiny \color{ForestGreen} $\left\lceil +16.3 \right\rceil$} & \underline{74.9} {\tiny \color{ForestGreen} +11.9} & 57.9 {\tiny \color{DarkRed} -5.5} & \underline{43.1} {\tiny \color{ForestGreen} +2.2} & \underline{64.2}  {\tiny \color{ForestGreen} $\left\lceil +25.0 \right\rceil$} & \underline{64.4} {\tiny \color{ForestGreen} $\left\lceil +21.3 \right\rceil$} & \underline{66.7} {\tiny \color{ForestGreen} +16.4} & \underline{65.0} {\tiny \color{ForestGreen} +0.0} & \underline{68.3} {\tiny \color{ForestGreen} +4.8}\\ 
		\bottomrule
	\end{tabular}
	\caption{\textbf{HTSS on pixel-labeled datasets with conflicting label spaces}. Performance of combined training (bottom rows) compared to single-dataset training (top rows). $L(\text{dataset})$ = 19, 20, 20, 33, 66 for the 5 datasets. $\left\lceil \text{+x} \right\rceil$: up to +x\%.}
	\label{tab:comb-labels-1}
\end{table*}

\vspace{7pt}
\noindent\textbf{Knowledgeability metric.}
\label{ssec:knowledgeability}
This metric quantifies in a single value the semantic richness of the output of a semantic segmentation system, by evaluating how many semantic concepts (atoms) it can recognize with sufficient segmentation accuracy. 
Using existing metrics, one way to achieve this is to report the size of the system's output label space and separately the IoU performance per class. However, this approach has some pitfalls. First, merely reporting the label space size is not a reliable metric for semantic richness of predictions, since the IoU performance for some classes can be very low or even zero. Second, IoU-based average aggregates,~\eg mIoU, do not reflect the number of recognizable classes, because they assess a system purely at segmentation level. Finally, these aggregates are intrinsically dependent on the size of the evaluated label space: as the size increases, the difficulty of assigning the correct class increases, eventually leading to mIoU reduction (due to the smoothing properties of averaging). The new metric is designed to explicitly consider the size of the label spaces of both the system output and the evaluated dataset together with the segmentation performance for the output classes.


The core of the metric is based on counting the number of classes that achieve an IoU higher than a threshold $t$ wrt. the total number of classes $c$ that are considered for computing the metric. To make the metric independent of the need for proper selection of $t$, the counting is averaged over a set of $N_T$ thresholds, which in this work are chosen to be equidistant,~\ie $T = \{0.0, ~1/N_T, ~\dots, ~1.0-1/N_T\}$.
Other values for T can be chosen depending on the application and datasets specifics. Assuming an output label space of a model that contains $L$ discreet semantic classes, the set of all per-class IoUs $\mathcal{E} = \{\text{IoU}_i\}_{i = 1}^L$ can be constructed by evaluating the model output against the ground truth. Subsequently, the set $\mathcal{E}$ is used to generate all the subsets $\tilde{\mathcal{E}}_t=\{\text{IoU}~|~\text{IoU} > t,~\text{IoU} \in \mathcal{E}\}$ containing the IoUs above the threshold $t$ from $T$. To this end, \textit{Knowledgeability} is defined as the $c$-normalized number of classes averaged over $T$:
\begin{equation}
	\mathcal{K}^c_T = \frac{1}{N_T} \sum_{t \in T} \frac{\min( | \tilde{\mathcal{E}}_t |, c )}{c} ~.
	\label{eq:knowledgeability-definition}
\end{equation}


This definition guarantees $\mathcal{K}^c_T$ to be between 0 and 1,~\ie $0 \le \mathcal{K} \le \mathcal{K}_\text{max} = \min\left(L, c\right) / c \le 1$, which is achieved by creating the sets $\tilde{\mathcal{E}}_t$ using the strictly greater condition ($\text{IoU} > t$) and by employing the $\min$ function. The bounds enable the use of the metric for comparison across datasets with different number of classes and semantic segmentation systems with different number of output classes. The new metric \textit{Knowledgeability} allows us to express the increase in the number of recognizable classes and at the same time consider the performance on these classes, without the need to choose a specific single threshold.


\subsection{Implementation details}
\label{ssec:setup}

\noindent \textbf{Convolutional network architecture.}~
The convolutional network follows the PSP architecture~\cite{zhao2017pyramid}, since it provides a good trade-off between segmentation performance, training time, and memory requirements. The backbone feature extractor is a ResNet-50~\cite{he2016deep} modified for segmentation by i) changing the block strides and dilation rates~\cite{yu2016dilated,zhao2017pyramid}, ii) projecting the output 2048-dimension features to 256-dimension features to reduce memory requirements, and iii) using a Pyramid Pooling Module~\cite{zhao2017pyramid}, as described in~\cite{meletis2018heterogeneous}.



\noindent\textbf{Hyper-parameter tuning and implementation details.}~
Two factors that emerge in multi-dataset training and that significantly affect the accuracy of CNNs for segmentation are the \textit{batch size} and the input \textit{image size}. The global and per-dataset \textit{batch size} are connected with the robustness of the optimization algorithm (SGD) and determine training balance across all classes. As the number of the output classes increase, bigger batch sizes are required. The second factor,~\ie the input \textit{image size} of the feature extractor, determines the scale and detail of the extracted features throughout the network. Ideally one would like to use high \textit{batch sizes} and \textit{image sizes} given the available (fixed) compute infrastructure. However, using more datasets for multi-dataset training forces using larger batch sizes, to guarantee that all dataset are represented in the batch. This in turn, requires reducing the \textit{image size} for satisfying of the available (fixed) compute infrastructure. Each of our experiments (subsections) involve training with different number of datasets and classes, thus we tune these two hyper-parameters separately per experiment to obtain optimal performance given the constraints. This leads to different baseline performance for each experiment and results can therefore not be compared across different experiments. But, within a single experiment, we guarantee that all methods achieve optimal results by finding the empirically best trade-off between \textit{batch size} and \textit{image size} given the memory constraints. The experiments are conducted on a machine with 4 Titan V GPUs.

\begin{table*}
	\centering
	\small
	\begin{tabular}{cc|cc|c||lll|lll}
		\toprule
		\multicolumn{4}{c}{Training datasets} & \multirow{3}{*}{\makecell{Output\\label\\space}} & \multicolumn{3}{c|}{Zero-shot generalization} & \multicolumn{3}{c}{Val-split generalization} \\
		\cmidrule(l{3pt}r{3pt}){6-8} \cmidrule(l{3pt}r{3pt}){9-11}
		\multicolumn{2}{c|}{pixel-labeled} & \multicolumn{2}{c|}{bbox-labeled} & & \multicolumn{3}{c|}{WildDash (unseen)} & \multicolumn{3}{c}{Cityscapes (seen)}\\
		Citys & CitysC & ECP & CitysC & & ECP-2 & CitysC-10 & All classes & ECP-2 & CitysC-10 & All classes\\ 
		\midrule
		\ding{51} & & & & C-20 & 12.9 & 16.7 & 23.0 & 65.0 & 69.6 & 61.3\\ 
		\ding{51} & \ding{51} & & & C-20 & 13.8 & 17.4 & 23.1 & 65.2 & 70.1 & 61.5\\ 
		\midrule
		\ding{51} & & \ding{51} &  & HTSS-20 & \underline{31.7} {\tiny \color{ForestGreen} $ +18.8 $} & 20.7 {\tiny \color{ForestGreen} $ +4.0 $} & 22.7 {\tiny \color{DarkRed} $ -0.3 $} & \underline{66.7} {\tiny \color{ForestGreen} $ +1.7 $} & 70.5 {\tiny \color{ForestGreen} $ +0.9 $} & 61.3 {\tiny \color{ForestGreen} $ +0.0 $}\\ 
		\ding{51} & & & \ding{51} & HTSS-20 & \underline{31.4} {\tiny \color{ForestGreen} $\left\lceil +17.6 \right\rceil$} & 20.8 {\tiny \color{ForestGreen} $\left\lceil +3.4 \right\rceil$} & \underline{24.4} {\tiny \color{ForestGreen} $\left\lceil +1.4 \right\rceil$} & 65.9 {\tiny \color{ForestGreen} $\left\lceil +0.9 \right\rceil$} & 70.5 {\tiny \color{ForestGreen} $\left\lceil +0.9 \right\rceil$} & \underline{63.4} {\tiny \color{ForestGreen} $\left\lceil +2.1 \right\rceil$} \\ 
		\ding{51} & & \ding{51} & \ding{51} & HTSS-20 & 31.1 {\tiny \color{ForestGreen} $\left\lceil +18.2 \right\rceil$} & \underline{26.2} {\tiny \color{ForestGreen} $\left\lceil +9.5 \right\rceil$} & 22.4 {\tiny \color{ForestGreen} $\left\lceil -0.6 \right\rceil$} & 64.0 {\tiny \color{DarkRed} $\left\lceil -1.0 \right\rceil$} & \underline{71.6} {\tiny \color{ForestGreen} $\left\lceil +2.0 \right\rceil$} & 61.7 {\tiny \color{ForestGreen} $\left\lceil +0.4 \right\rceil$} \\ 
		\bottomrule
	\end{tabular}
	\caption{\textbf{HTSS on pixel-labeled and bounding-box-labeled datasets with non-conflicting label spaces}. Accuracy [mIoU \%] on seen and unseen datasets for the specific class subsets (ECP-2, CitysC-10) that receive the extra weak supervision from ECP and CitysC and all classes. The pixel-labeled CitysC (row 2) is used to set the oracle for the experiments involving the weakly-labeled CitysC (rows 4, 5). C-20 and HTSS-20 label spaces coincide, the HTSS prefix is used to emphasize mixed-supervision training using the HTSS framework. $\left\lceil \text{+x} \right\rceil$: up to +x\%.}
	\label{tab:combine-annotations}
\end{table*}

\subsection{Strong supervision, conflicting label spaces}
\label{ssec:exps-combine-label-spaces}
In the first set of experiments (Tab.~\ref{tab:comb-labels-1}), we focus on combining pixel-labeled datasets with conflicting label spaces (case of third row in Tab.~\ref{tab:overview-methods}). This scenario is commonly occurring, where different pixel-labeled datasets (for semantic segmentation) are annotated at conflicting levels of semantic granularity. In the experiments, the label spaces of three datasets (Cityscapes, Vistas, IDD) are combined, as described in Sec.~\ref{sec:label-taxonomy}, resulting in the taxonomy of Fig.~\ref{fig:hierarchy} (without the MTS dataset). The direct solution of training with the union of datasets and their label spaces is not applicable, since the semantic conflicts among the label spaces introduce ambiguities in the concatenated output label space. For example, the \textit{rider} Cityscapes class conflicts with the \textit{motorcyclist} and \textit{bicyclist} Vistas classes. These conflicts are resolved by generating a universal taxonomy (Sec.~\ref{sec:label-taxonomy}).

We train on all combinations of three pixel-labeled datasets using the HTSS methodology after solving the conflicts in their label spaces. We compare the accuracy and knowledgeability (Sec.~\ref{ssec:metrics}) between HTSS networks against single-dataset trainings. For these experiments, the input image size is $799 \times 799$ and batch size formation is 1 image from Cityscapes, 2 from Vistas and 1 from IDD, for each GPU.

\noindent\textit{Accuracy results}. As can be seen in Tab.~\ref{tab:comb-labels-1}, the HTSS-68 network trained on all datasets outperforms 6 out of 7 other networks for seen (columns 3-5) or unseen datasets (columns 1-2). The improvements are due to training a single model with diverse images from multiple datasets (9 times more than Cityscapes and 2 times more than Vistas), which is possible after solving conflicts in the label spaces. An exception is the case of the IDD val split, where the single-dataset training matches or outperforms HTSS networks. After careful visual examination of the dataset, we have observed that the semantic annotations of IDD have a high degree of overlapping concepts. For example, the \textit{road} class and the \textit{drivable fallback} class have partially overlapping definitions,~\ie they both contain semantic atoms like \textit{pothole} or \textit{crosswalk}. This contradicts our hypothesis on assuming non-conflicting semantic class definitions (refer to Eq.~\eqref{eq:hypothesis-1}) and possibly explains the discrepancy in the results.

\noindent\textit{Knowledgeability results.} The HTSS networks are able to segment, in a single pass, an image over more semantic concepts with high attainable mIoU, as demonstrated by \textit{Knowledgeability}. Moreover, this is proportional to the size of the output label space (\eg columns 6, 7, 10).

\subsection{Strong \& weak supervision, non-conflicting label spaces}
\label{ssec:exps-strong-weak-nonconfl}
This section explores HTSS on a mix of strongly- and weakly-labeled datasets that have non-conflicting label spaces. The hypothesis under investigation is that a very large quantity of weak labels (exponentially more samples but with poor localization) can be beneficial if used in combination with strong labels, especially for under-represented classes.

Using the approach developed in Sec.~\ref{sec:weak-superv} we train HTSS networks on the pixel-labeled Cityscapes and Cityscapes Coarse, the bounding-box-labeled ECP, and the bounding-box-labeled Cityscapes Coarse, that we generated from the pixel labels (see Tab.~\ref{tab:datasets-overview}). The label space used is the common 20 classes for Cityscapes (C-20), which we call also HTSS-20, as it is a subset of the complete taxonomy of Fig.~\ref{fig:hierarchy}. The results are provided in Tab.~\ref{tab:combine-annotations}. For these experiments, the input image size is $699 \times 699$ and batch size formation is 1 image from Cityscapes, 2 from Cityscapes Coarse, and 2 from ECP, for each GPU.

There are two interesting comparisons to discuss. First, training with Cityscapes and Cityscapes Coarse irrespective of the supervision type (rows 2 and 4) should in principle give similar performance since the supervision information is the same. However the HTSS network shows higher accuracy for all class subsets either in the same or different image domains. Specifically, the two underrepresented classes of Cityscapes (\textit{person} and \textit{rider},~\ie ECP-2) have the higher difference in accuracy by +18.8\% for Wild Dash. This indicates that the generated pixel pseudo-labels by HTSS (see Fig.~\ref{fig:temp-loss-comps}) provide more complete training cues than the original pixel labels of Cityscapes Coarse. The second comparison (rows 1 and 3) highlights that bounding-box labels from ECP (different image domain) increase the generalization accuracy compared to 

From the results of Tab.~\ref{tab:combine-annotations} we conclude that the HTSS framework can successfully leverage weak supervision to increase segmentation accuracy on selected classes (\eg vulnerable road users,~\ie ECP-2) and at the same time maintain or slightly improve the accuracy achieved by strong supervision (Cityscapes).

A second experiment examines the segmentation accuracy of specific classes when adding weak supervision from bounding boxes and image tags. The results are provided in Tab.~\ref{tab:perf-detail-citys}) and demonstrate that an increasing amount of weak supervision improves the segmentation accuracy accordingly. Using weak supervision from the Open Images dataset (rows 2 and 3) improves the average mIoU accuracy up to +2.9\%, while specific classes are substantially benefited with an increase of up to +13.2\% IoU. The inclusion of image-tag supervision, improves or maintains IoUs for 6 out of 8 classes, however the improvement is less significant compared to bounding-box supervision only. This shows that weaker and less localized forms of supervision have smaller gains in segmentation performance.

\begin{table}
	\centering
	\footnotesize
	\setlength\tabcolsep{2.2pt}
	\begin{tabular}{@{}ccc|cccccc|c||cc|c@{}}
		\toprule
		&&& \multicolumn{10}{c}{Val-split generalization - Cityscapes (seen)} \\
		\multicolumn{3}{c|}{Train datasets} & \mrowrot{2}{Bicycle} & \mrowrot{2}{Bus~~} & \mrowrot{2}{Car~~} & \mrowrot{2}{Motorc.} & \mrowrot{2}{Train~} & \mrowrot{2}{Truck} & \mrowcellrot{2}{l}{Vehicle\\mIoU} & \mrowrot{2}{Person} & \mrowrot{2}{Rider} & \mrowcellrot{2}{l}{Human\\mIoU}\\
		Citys & \multicolumn{2}{c|}{Open Im.} & & & & & & & & & & \\
		pixel & bbox & tag & & & & & & & & & & \\
		\midrule
		\ding{51} & & & 67.8 & 80.1 & 92.3 & \underline{51.9} & \underline{69.6} & 63.2 & 70.8 & 70.9 & 48.5 & 59.7 \\
		\ding{51} & \ding{51} & & 68.7 & \underline{82.1} & \underline{92.9} & 50.2 & \underline{69.8} & 71.9 & \underline{72.6} & 72.5 & 51.2 & 61.9 \\
		\ding{51} & \ding{51} & \ding{51} & \underline{69.1} & 79.7 & \underline{92.8} & 48.9 & \underline{69.6} & \underline{76.3} & \underline{72.7} & \underline{72.9} & \underline{52.5} & \underline{62.7}\\
		\bottomrule
	\end{tabular}
	\caption{\textbf{HTSS on pixel-, bounding-box-, and image-tag- labeled images with non-conflicting label spaces}. Accuracies (mIoU \%) for the specific classes that receive extra supervision from the weakly-labeled Open Images dataset.}
	\label{tab:perf-detail-citys}
\end{table}

\begin{table}
	\centering
	\small
	\setlength\tabcolsep{2.5pt}
	\begin{tabular}{cc|cc|c||lllll}
		\toprule
		\multicolumn{4}{c}{Train datasets} \\
		\multicolumn{2}{c|}{pixel} & \multicolumn{2}{c|}{bbox} \\
		\mrowrot{3}{~~Citys} & \mrowrot{3}{~~~CitysT} & \mrowrot{3}{~~~CitysT} & \mrowrot{3}{~~MTS} & \multirow{3}{*}{\makecell{Output\\label\\space}} & \multicolumn{1}{c}{WildDash} & \multicolumn{4}{c}{Cityscapes T. Signs (seen)}\\
		\cmidrule(l{3pt}r{3pt}){6-6} \cmidrule(l{3pt}r{3pt}){7-10}
		& & & & & W-19 & CitysT-14 &C-20 & \multicolumn{2}{c}{CitysT-34}\\
		& & & & & mIoU & mIoU & mIoU & mIoU & $\mathcal{K}^{34}$\\
		\midrule
		\ding{51} & & & & C-20 & 27.1 & n/a & 70.3 & n/a & 39.3\\
		\ding{51} & \ding{51} & & & CT-34 & 27.8 & 17.7 & 69.5 & 47.5 & 46.2\\
		\ding{51} & & \ding{51} & & HTSS-34 & 30.2 {\tiny \color{ForestGreen} $\left\lceil +3.1 \right\rceil$} & 17.0 & 69.8 {\tiny \color{ForestGreen} $\left\lceil +0.3 \right\rceil$} & 46.9 & 44.5 {\tiny \color{ForestGreen} $\left\lceil +5.2 \right\rceil$}\\
		\midrule
		\ding{51} & & & \ding{51} & HTSS-70 & 28.9 & 11.6 & 70.7 & 45.6 & 44.3\\
		\bottomrule
	\end{tabular}
	\caption{\textbf{HTSS on pixel-labeled and bounding-box-labeled datasets with conflicting label spaces.} Performance on seen and unseen datasets for all classes (W-19, CitysT-34) and for class subsets (C-20, CitysT-14). Cityscapes Traffic Signs (CitysT) is originally a pixel-labeled dataset, which we convert to bounding boxes~\cite{meletis2019boosting} (bbox column). CitysT-34 = C-20 $\cup$ CitysT-14, HTSS-70 = C-20 $\cup$ MTS-50.
	}
	\label{tab:combine-annotations-conflicting-1}
\end{table}

\subsection{Strong \& weak supervision, conflicting label spaces}
\label{ssec:exps-combine-supervision}
The most challenging scenario involves multi-dataset training with mixed supervision and conflicting label spaces. To investigate this scenario, we augment the label space of Cityscapes with traffic sign classes from the bounding-box-labeled datasets Cityscapes Traffic Signs (14 t. signs) and MTS (50 t. signs). In this case, the methodology developed in Sec.~\ref{ssec:general-case} is employed, the HTSS network is trained with input image size of $599 \times 599$, and the batch size arrangement is 1 image from Cityscapes, 3 from Cityscapes Traffic Signs, and 3 from Mapillary Traffic Signs, for each GPU.

Table~\ref{tab:combine-annotations-conflicting-1} shows the results. We evaluate all networks on the original Cityscapes classes (C-20), the traffic sign classes subset (CitysT-14), all CitysT-34 classes (CitysT-14 traffic signs + C-20), as well as, on the unseen Wild Dash, whose classes are a subset of Cityscapes (W-19 $\subset$ \mbox{C-20}). Comparing row 3 (HTSS-34) with the first two rows we observe that using the HTSS methodology the network is able to learn from weak supervision and even surpass the fully pixel-level supervised oracle (row 2) in some cases. Moreover, the results of the last row indicate that using MTS, which has a different image domain than Cityscapes, the network is able to segment Cityscapes traffic signs to some extent (11.6\%). Overall, these experiments demonstrate the ability to: i) learn from mixed supervision, ii) learn from the same or different image domains, and iii) maintain or slightly improve generalization accuracy for the pixel-labeled classes (columns 1, 3).

\begin{figure}
	\begin{subfigure}{0.8\linewidth}
		\includegraphics[width=\linewidth]{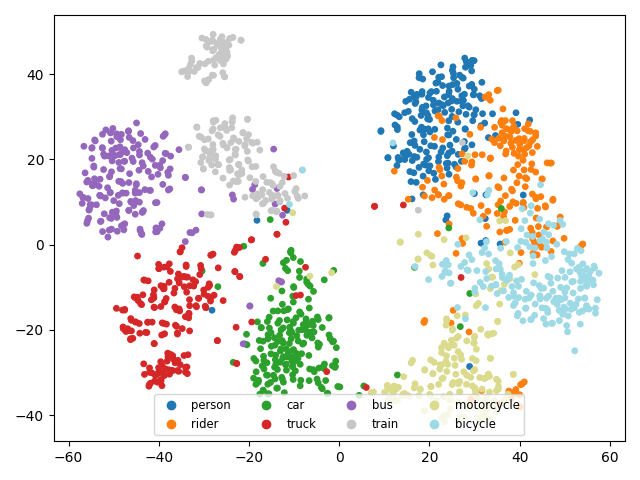} 
		\caption{Single-dataset training on Cityscapes.\\~}
		\label{fig:subim1}
	\end{subfigure} \\
	\begin{subfigure}{0.8\linewidth}
	    \includegraphics[width=\linewidth]{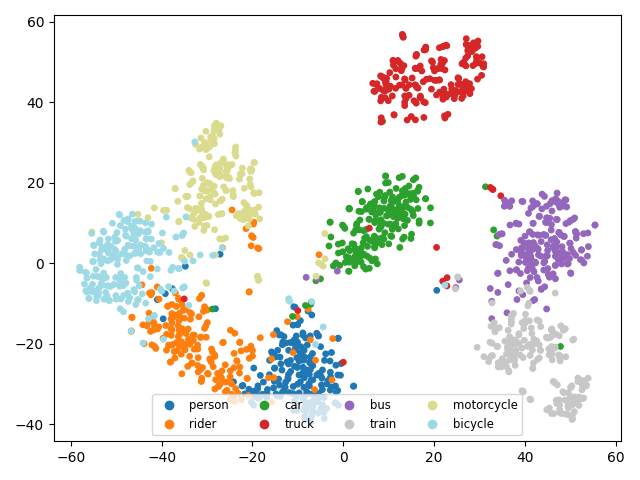}
		\caption{HTSS on Cityscapes + Vistas + IDD.\\~}
		\label{fig:subim2}
	\end{subfigure} ~
	\caption{Progress of features, while adding more datasets, as a 2-D t-SNE visualization, when using same t-SNE hyper-parameters. Clusters become less scattered (intra-class distance) and better separated (inter-class distance). 
	}
	\label{fig:tsne}
\end{figure}

\begin{table}
	\centering
	\footnotesize
	\setlength\tabcolsep{4.5pt}
	\begin{tabular}{ll|cccc}
		\toprule
		Model & \makecell{Conflicts\\resolution} & \makecell{Memory\\$\Delta$ Params} & \makecell{Inference\\$\Delta$ ms} & \makecell{Unseen dts.\\mmIoU} & $\mathcal{K}^{66}$\\
		\midrule
		\makecell[l]{one per\\dataset} & \makecell[l]{post-proc.\\merging} & $+ 5.4 \cdot 10^7$ & $+ 127.1$ & 46.2 & 62.6\\ 
		\midrule
		\multirow{2}{*}[2pt]{\makecell[l]{shared\\backbone,\\head per\\dataset}} & \makecell[l]{common\\classes} & $+ 2.1 \cdot 10^5$ & $+ 1.7$ & 34.2 & 45.6\\
		\cmidrule{2-6}
		& \makecell[l]{post-proc.\\merging} & $+ 2.1 \cdot 10^5$ & $+ 1.8$ & 45.4 & 62.8\\ 
		\midrule
		\multirow{2}{*}[2pt]{\makecell[l]{shared\\backbone,\\single\\head}} & \makecell[l]{common\\classes} & \textit{reference} & \textit{reference} & 30.2 & 42.1\\ 
		\cmidrule{2-6}
		& \makecell[l]{\textbf{semantic}\\\textbf{atoms (ours)}} & $+ 5.1 \cdot 10^3$ & $+ 0.0$ & 50.5 & 68.3 \\ 
		\bottomrule
	\end{tabular}
	\caption{Common baselines methodologies for combining three pixel-labeled datasets (Cityscapes, Vistas, IDD) with conflicting label spaces. All methods use ResNet-50 backbones and softmax classifiers. The $\Delta$'s for the total number of parameters (Params) and the single-image inference time (ms) are w.r.t. the \textit{reference} row 4,~\ie keeping only the common, non-conflicting classes from all datasets.}
	\label{tab:comb-labels-2}
\end{table}

\subsection{Ablations and Insights}
\label{ssec:exps-ablations}
An analysis is provided for the conducted experiments. First, an ablation on how the amount of weak supervision affects performance is presented in Tab.~\ref{tab:size-matters} for the experiment of Sec.~\ref{ssec:exps-combine-supervision}. An increasing number of images and bounding boxes from the weakly-labeled dataset are added per step. We observe that as weakly-labeled images are included, the segmentation performance increases accordingly.

Second, we provide t-SNE plots in Fig.~\ref{fig:tsne} for experiments from Tab.~\ref{tab:comb-labels-1}.
The plots capture the 2-D projections of the output of feature extractor before and after adding multiple datasets. It can be seen that the representations have better properties from the perspective of classificability/separability.

Finally, we provide comparisons of the HTSS methodology against various baselines in Tab.~\ref{tab:comb-labels-2} examining memory and time factors wrt. the attained performance. The first three rows describe direct solutions using existing trained networks and post processing for solving conflicts. The single-network approach (fourth row) is the closest to HTSS, but resolves conflicts by maintaining only the common classes. This leads to a significant loss in \textit{Knowledgeability}, as the number of recognizable classes reduce. Overall, the HTSS approach uses a reduced number of parameters and performs fast inference, since it uses a common backbone and a single classifier.

\begin{table}
	\centering
	\setlength\tabcolsep{4.0pt}
	\begin{tabular}{cc||cc}
		\toprule
		pixel & bbox-labeled & \multicolumn{2}{c}{Cityscapes (C-20)}\\
		Citys & Open Images & mAcc & mIoU \\
		\midrule
		\ding{51} & - & 81.2 & 70.2\\
		\midrule
		\ding{51} & $1k$ images ($17.3k$ bboxes) & 80.7 & 69.8\\ 
		\ding{51} & $10k$ images ($140.4k$ bboxes) & 81.6 & 70.6\\ 
		\ding{51} & $100k$ images ($1185.8k$ bboxes) & 83.7 & 72.3\\ 
		\bottomrule
	\end{tabular}
	\caption{Segmentation accuracy with different number of bounding boxes used to generate pseudo labels from the weakly-labeled Open Images.}
	\label{tab:size-matters}
\end{table}

\section{Conclusion}
We presented the HTSS framework for simultaneously training FCNs on multiple heterogeneous datasets for semantic segmentation. We explored heterogeneous training with various combinations of strongly (pixel) and weakly (bounding boxes, image tags) labeled datasets. The experiments showed that HTSS improved, in the majority of the combinations, three aspects of the predicted results: i) the segmentation performance on test splits of seen (training) datasets, generalization on unseen datasets, and awareness of semantic concepts, expressed by the proposed \textit{Knowledgeability} metric. Our HTSS framework does not require any extra labeling for weakly-labeled datasets, while the only manual step consists of defining a semantic taxonomy of the label spaces of the employed datasets, solely in the case of conflicting label spaces. HTSS allows supplementing the pixel-labeled training data with other relevant datasets that otherwise would not be compatible. These properties make the HTSS approach useful to many applications in which training data for semantic segmentation are too scarce to achieve required performance.



%








\bibliographystyle{IEEEtran}
\bibliography{biblio}
%

%

\vspace{200pt}

\begin{IEEEbiographynophoto}{Panagiotis Meletis}
Panagiotis Meletis holds a PhD from the Eindhoven University of Technology. He works on advancing deep learning techniques for autonomous vehicles within the Mobile Perception Systems lab. His research involves the design of novel algorithms towards holistic visual scene understanding. Panagiotis received his MSc in Electrical and Computer Engineering from the National Technical University of Athens in 2015. His current research interests include deep learning techniques for machine perception, automated visual reasoning, and explainable and sustainable artificial intelligence.
\end{IEEEbiographynophoto}


\begin{IEEEbiographynophoto}{Gijs Dubbelman}
PhD, is an associate professor with the Eindhoven University of Technology and heading the Mobile Perception Research (MPS) lab, which aims to improve the real-time perception capabilities of mobile sensor platforms through research on artificial intelligence. Key research areas of the MPS lab are 3-D computer vision, multi-modal pattern recognition, and deep learning. Formerly, Gijs Dubbelman was a member of the Field Robotics Center of Carnegie Mellon’s Robotics Institute, where he performed research on Visual-SLAM for autonomous robots and vehicles.
\end{IEEEbiographynophoto}






\vfill

\end{document}